\def\BuildCombinedPaper{1}
\documentclass{article}

\PassOptionsToPackage{sort}{natbib}

\usepackage{iclr2027_conference,times}

\usepackage{amsmath,amsfonts,bm}

\def\eqref#1{equation~\ref{#1}}

\def\1{\bm{1}}

\DeclareMathAlphabet{\mathsfit}{\encodingdefault}{\sfdefault}{m}{sl}
\SetMathAlphabet{\mathsfit}{bold}{\encodingdefault}{\sfdefault}{bx}{n}

\usepackage{amsmath}
\usepackage{amssymb}
\usepackage{mathtools}
\usepackage{booktabs}
\usepackage{multirow}
\usepackage{graphicx}
\usepackage{subcaption}
\usepackage{xcolor}
\usepackage{colortbl}
\usepackage{microtype}
\usepackage{url}
\usepackage{hyperref}
\usepackage{wrapfig}
\usepackage{float}

\setcitestyle{numbers,square,comma}

\usepackage[capitalise,noabbrev,nameinlink]{cleveref}
\crefname{section}{Sec.}{Secs.}
\Crefname{section}{Sec.}{Secs.}
\crefname{subsection}{Sec.}{Secs.}
\Crefname{subsection}{Sec.}{Secs.}
\crefname{subsubsection}{Sec.}{Secs.}
\Crefname{subsubsection}{Sec.}{Secs.}
\crefname{figure}{Fig.}{Figs.}
\Crefname{figure}{Fig.}{Figs.}
\crefname{equation}{Eq.}{Eqs.}
\Crefname{equation}{Eq.}{Eqs.}
\crefname{table}{Tab.}{Tabs.}
\Crefname{table}{Tab.}{Tabs.}

\definecolor{linkblue}{rgb}{0.10,0.20,0.55}
\definecolor{tableRankBase}{HTML}{555FFF}
\colorlet{tableRank1}{tableRankBase!32}
\colorlet{tableRank2}{tableRankBase!20}
\colorlet{tableRank3}{tableRankBase!10}

\hypersetup{
  colorlinks=true,
  linkcolor=linkblue,
  citecolor=linkblue,
  urlcolor=linkblue
}

\graphicspath{{figures/}}

\newcommand{\method}{\textsc{Mix Forcing}}
\newcommand{\moba}{\textsc{MoBA}}

\newcommand{\rankfirst}[1]{\cellcolor{tableRank1}\textbf{#1}}
\newcommand{\ranksecond}[1]{\cellcolor{tableRank2}#1}
\newcommand{\rankthird}[1]{\cellcolor{tableRank3}#1}

\usepackage{array}
\usepackage{tabularx}
\newcolumntype{N}{>{\centering\arraybackslash}X}
\usepackage{flafter}
\usepackage{placeins}
\usepackage{needspace}

\AddToHook{env/figure/begin}[main-layout]{%
  \setlength{\abovecaptionskip}{4pt}%
  \setlength{\belowcaptionskip}{-4pt}}
\AddToHook{env/figure*/begin}[main-layout]{%
  \setlength{\abovecaptionskip}{4pt}%
  \setlength{\belowcaptionskip}{-4pt}}
\AddToHook{env/wrapfigure/begin}[main-layout]{%
  \setlength{\abovecaptionskip}{4pt}%
  \setlength{\belowcaptionskip}{-4pt}}

\AddToHook{env/table/begin}[main-layout]{%
  \setlength{\abovecaptionskip}{0pt}%
  \setlength{\belowcaptionskip}{3pt}}
\AddToHook{env/table*/begin}[main-layout]{%
  \setlength{\abovecaptionskip}{0pt}%
  \setlength{\belowcaptionskip}{3pt}}

\AddToHook{env/table/end}[main-layout]{\par\vspace{-4pt}}
\AddToHook{env/table*/end}[main-layout]{\par\vspace{-4pt}}

\input{public_style}
\ifdefined\BuildCombinedPaper
  \input{combined_layout}
  \usepackage{titletoc}
\newcommand{\supplementarycontents}{%
  \startcontents[appendix]%
  \section*{Supplementary Contents}%
  \begingroup
    \hypersetup{linktoc=all}%
    \printcontents[appendix]{}{1}{\setcounter{tocdepth}{2}}%
  \endgroup
  \clearpage
}

\fi

\title{Mix Forcing: Causal Pre-Training via Mixture Attention for Interactive World Models}

\author{%
  {\fontfamily{EBGaramond-LF}\selectfont
  \mbox{Zelin Gao$^{1}$}, \mbox{Hao Ouyang$^{2,*}$},
  \mbox{Jiapeng Zhu$^{2}$}, \mbox{Qiuyu Wang$^{2}$}\\ \mbox{Jingye Chen$^{1}$},
  \mbox{Zichen Liu$^{1}$}, \mbox{Xing Zhu$^{2}$}, \mbox{Yujun Shen$^{2}$},
  \mbox{Yinghao Xu$^{1,\dagger}$}\\[2pt]
  {\small $^{1}$The Hong Kong University of Science and Technology\quad $^{2}$Robbyant}\\
  {\small $^{*}$Project Lead\quad $^{\dagger}$Corresponding Author}}%
}
\hypersetup{
  pdfauthor={Zelin Gao, Hao Ouyang, Jiapeng Zhu, Qiuyu Wang, Jingye Chen, Zichen Liu, Xing Zhu, Yujun Shen, Yinghao Xu},
  pdftitle={Mix Forcing: Causal Pre-Training via Mixture Attention for Interactive World Models}
}

\begin{document}

\maketitle

\begin{abstract}
While interactive video world models require autoregressive generation, directly adapting pretrained bidirectional video diffusion models into autoregressive ones with teacher forcing can disrupt their visual priors and degrade autoregressive generation quality.
In this paper, we introduce \method{}, a causal pre-training paradigm for interactive video world models, implemented with the Mixture-of-Bidirectional-and-Autoregressive~(\moba{}) attention.
\method{} jointly trains a video diffusion model with both teacher forcing and bidirectional training, yielding a strong causal initialization for few-step post-training.
In self-attention, \method{} leverages teacher forcing to learn causal next-chunk prediction for autoregressive rollout, while using bidirectional attention to preserve pretrained visual priors.
To support online text interaction, \method{} further adopts a chunk-wise cross-attention design that conditions each video chunk on a shared environment prompt and its own local prompt, without exposing future instructions to the current chunk.
Following causal pre-training on a 14B image-to-video model, we apply consistency and distribution matching distillation to obtain a four-step autoregressive generator.
Experiments on our internal benchmark and public benchmarks, including WBench and SANA-WM-Bench, show that our model achieves state-of-the-art visual quality and timely event following with precise camera control.
Notably, our model supports high-fidelity long-horizon generation without quality degradation even over one-hour rollouts.
\par\medskip
{\centering\small\upshape\bfseries
Project Page: \url{https://zelingao98.github.io/mixforcing}\par}
\end{abstract}
\begin{figure}[h!]
\centering
\includegraphics[width=\linewidth]{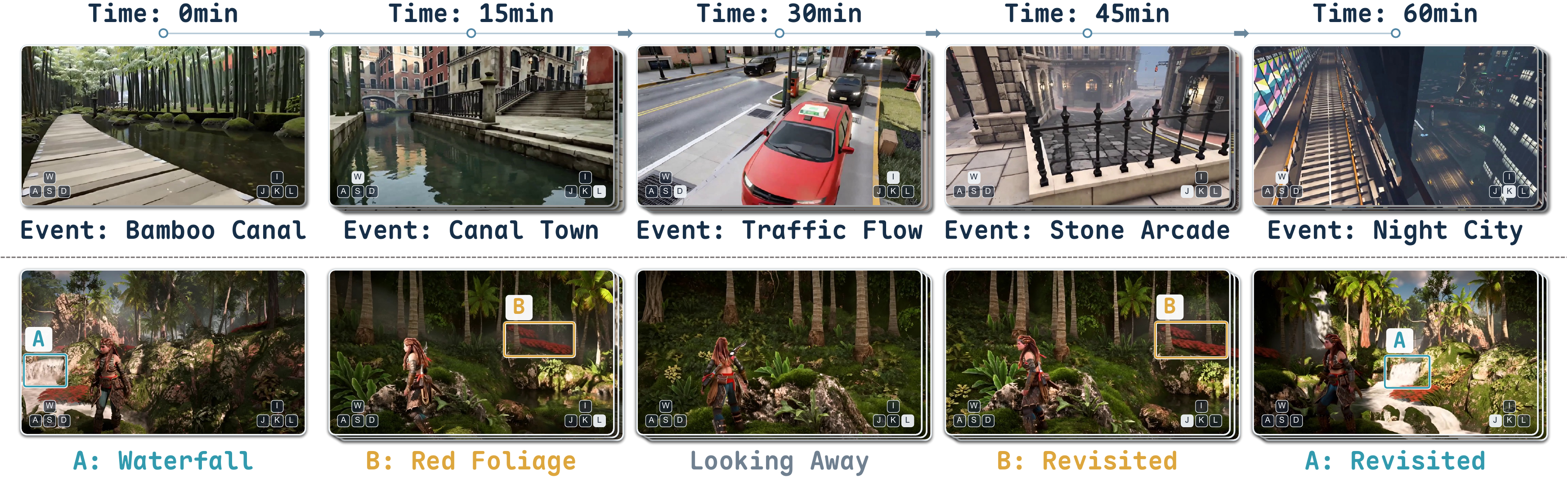}%
\caption{\textbf{Long-horizon interactive video rollout with camera control and memory.}
Top: a single one-hour autoregressive rollout generated by our model trained with \method{}, following successive prompt updates under keyboard camera control.
Bottom: a memory case, where the camera looks away and later returns, and previously observed regions~(A: waterfall; B: red foliage) remain consistent when revisited.
}
\label{fig:teaser}
\end{figure}

\section{Introduction}
\label{sec:introduction}
Interactive video world models~\cite{li2026causal,team2026advancing,gao2026infinite,zhang2026echowm,zhu2026sanawm,sun2025worldplay,he2025matrixgame2,zhang2026native}, which predict future states conditioned on historical context and user inputs, have shown great potential for world simulation, video gaming, and embodied intelligence.
Since users act based on past observations, each video chunk must be generated without access to future frames or future instructions, which calls for causal autoregressive generation.

Recent works~\cite{team2026advancing,zhu2026sanawm,he2025matrixgame2} build upon pretrained video diffusion models to leverage their strong visual priors.
However, these priors are learned for bidirectional generation, where all frames are jointly denoised with access to both past and future frames.
Some methods~\cite{huang2026self,yin2025slow} directly distill bidirectional models into few-step causal generators via ordinary differential equation~(ODE) initialization and distribution matching distillation~(DMD), but the trajectories of the bidirectional teacher depend on future information, preventing the causal student from converging to the target flow map.
To address this, other methods~\cite{zhu2026causal,zhao2026causal} first adapt pretrained models into many-step causal models with teacher forcing before few-step distillation, making this causal pre-training stage the foundation of the final generator.

However, teacher forcing departs from the pretraining setting of bidirectional video diffusion models in two aspects.
In attention, each chunk attends only to its history rather than the full video; in noise level, each noisy chunk is predicted from clean historical context rather than jointly denoised with other chunks at a shared noise level, which encourages velocity prediction to over-rely on clean context.
Both deviations disrupt pretrained visual priors and degrade autoregressive generation quality.
Moreover, causal pre-training methods typically condition the whole video on a single global prompt, which exposes future events to the current chunk.
Streaming methods ~\cite{yang2025longlive} support prompt updates by recaching the history under each new prompt, which rewrites the context of past frames.

In this paper, we introduce \textbf{\method{}}, a causal pre-training paradigm for interactive video world models, implemented with the \textbf{Mixture-of-Bidirectional-and-Autoregressive~(\moba{})} attention block.
The key idea of \method{} is to mix teacher forcing and bidirectional training: teacher forcing learns causal next-chunk prediction for autoregressive rollout, while bidirectional training retains the pretraining setting in both attention and noise level, preserving pretrained visual priors during causal adaptation.
In addition, \method{} conditions each teacher forcing chunk on a shared environment prompt and its own local prompt, enabling online text interaction without exposing future instructions to the current chunk or recomputing the KV cache of past chunks.

\begin{figure}[!t]
\centering
\includegraphics[width=0.95\textwidth]{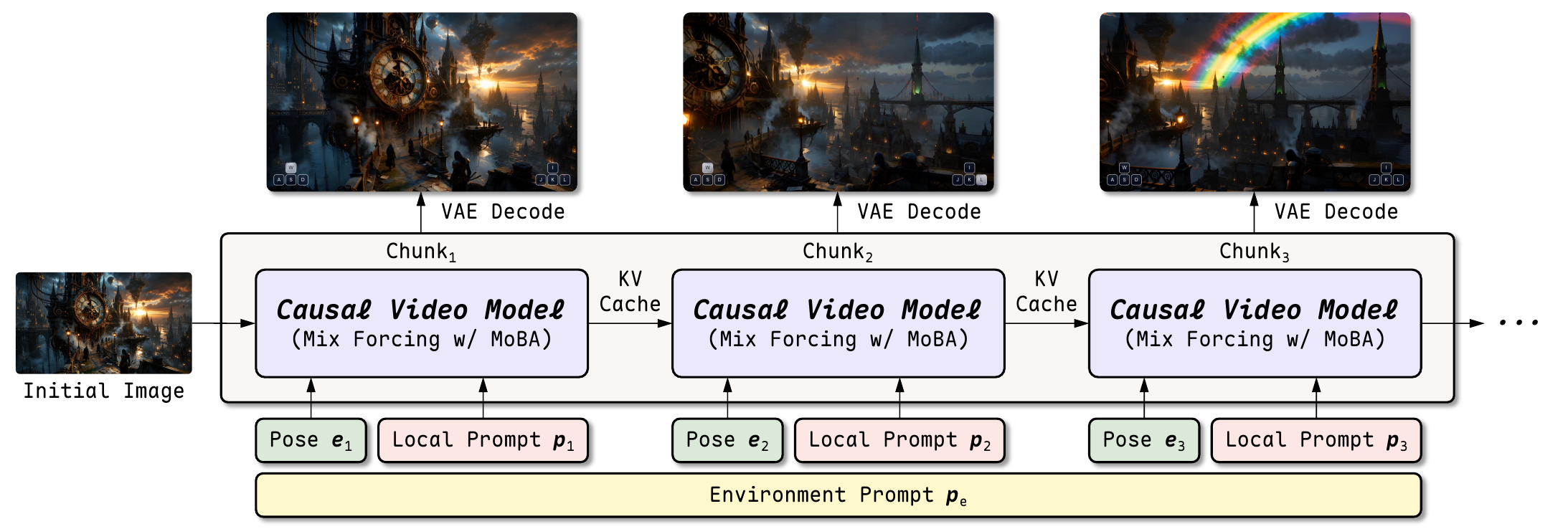}%
\caption{\textbf{Pipeline of our interactive video world model.}
Given an initial image and an environment prompt $\mathbf{p}_{e}$, an interactive world is simulated by our four-step causal video model.
Based on the KV-cached history context and user inputs, including the camera pose $\mathbf{e}_{i}$ and the local prompt $\mathbf{p}_{i}$, each latent chunk is generated autoregressively and decoded into a video clip by the VAE decoder.
}
\label{fig:pipeline}
\end{figure}

Based on \method{}, we build an interactive video world model with camera control and online text interaction~(\cref{fig:teaser}).
Starting from a 14B image-to-video model, we perform causal pre-training with \method{}, and then apply consistency distillation~(CD)~\cite{song2023consistency} and DMD to obtain a four-step generator for KV-cached autoregressive rollout~(\cref{fig:pipeline}).
Experiments on our internal benchmark and public benchmarks show that our model achieves state-of-the-art visual quality and timely event following with precise camera control, and sustains high-fidelity generation over one-hour rollouts.
Ablation studies further show that \method{} provides a stronger initialization for few-step post-training than teacher forcing alone.

Our contributions are summarized as follows:
\begin{itemize}
\item We propose the \moba{} attention block, which mixes teacher forcing and bidirectional training to preserve pretrained visual priors during causal adaptation, with chunk-wise prompt conditioning for online text interaction.
\item We build an interactive video world model through causal pre-training with \method{} and few-step post-training with CD and DMD, supporting camera control, online text interaction, and four-step KV-cached autoregressive rollout.
\item Experiments on internal and public benchmarks show that our model achieves state-of-the-art visual quality and timely event following with precise camera control.
\end{itemize}

\section{Related Work}
\label{sec:related-work}
\paragraph{Interactive and Controllable Video World Models.}
Interactive world models predict visual states conditioned on user actions~\cite{bruce2024genie,alonso2024diffusion,valevski2025diffusion}, while controllable video generation also supports text instructions and camera trajectories.
Prior work incorporates camera geometry through ray representations~\cite{he2024cameractrl,he2025cameractrl,zheng2024cami2v}, trajectory embeddings~\cite{wang2024motionctrl,bai2025recammaster}, or relative positional encodings~\cite{li2026cameras}, enabling geometry-aware viewpoint control in video generation.
Vid2World~\cite{huang2025vid2world} adapts pretrained video diffusion for autoregressive action control, while LongLive~\cite{yang2025longlive} supports streaming prompt updates by recaching the history under each new prompt.
MotionStream~\cite{shin2026motionstream} supports interactive motion control during streaming video generation.
Other video world models support interactive or long-horizon generation~\cite{yu2025gamefactorly,che2025gamegen,he2025matrixgame2,sun2025worldplay,team2026advancing,zhu2026sanawm,gao2026infinite,zhang2026echowm}.
In this work, we aim to use causal pre-training for interactive video world models, maintaining the visual priors of pretrained video diffusion models while enabling online text interaction through chunk-wise conditioning.
\paragraph{Causal Autoregressive Video Generation.}
Autoregressive video models generate frames or chunks sequentially~\cite{deng2025autoregressive,teng2025magi}, with some using KV-cached streaming inference~\cite{shazeer2019fast,li2026causal}.
FramePack~\cite{zhang2026frame} compresses historical frames for long rollouts, while diffusion-forcing approaches assign varying noise levels across video frames~\cite{chen2024diffusion,xie2025progressive,chen2025skyreels}.
To adapt pretrained bidirectional video diffusion models for online video rollout, prior methods initialize few-step causal students from probability-flow ODE trajectories~\cite{song2020score,karras2022elucidating,yin2025slow} and subsequently improve them through distribution matching distillation~\cite{yin2024one}.
Self Forcing~\cite{huang2026self} addresses rollout errors using self-generated history; Self-Forcing++~\cite{cui2026self} and Self Gradient Forcing~\cite{zhuang2026self} extend this approach through teacher supervision on long self-generated rollouts and context-gradient reconstruction, respectively.
Resampling Forcing~\cite{guo2026end} and Rolling Forcing~\cite{liu2026rolling} use self-resampling and joint denoising, respectively.
Causal Forcing~\cite{zhu2026causal} and Causal Forcing++~\cite{zhao2026causal} improve few-step distillation using a causal teacher and causal consistency distillation, respectively.
Reward Forcing~\cite{lu2026reward} uses reward-weighted distribution matching distillation for streaming video generation.
In \method{}, we jointly optimize teacher forcing and bidirectional attention during causal pre-training to retain pretrained visual priors before few-step post-training.

\FloatBarrier
\section{Methodology}
\label{sec:method}
In this section, we present \method{}, a causal pre-training paradigm for interactive video world models.
We first review flow matching and teacher forcing for interactive video modeling~(\cref{sec:prelim}).
We then introduce the \moba{} attention block that implements \method{}, which mixes teacher forcing and bidirectional training to preserve pretrained visual priors, with chunk-wise prompt conditioning for online text interaction~(\cref{sec:moba}).
Finally, as shown in \cref{fig:pipeline}, we apply \method{} to build an interactive video world model with camera control and online text interaction~(\cref{sec:ivwm}).

\subsection{Preliminary}
\label{sec:prelim}
\paragraph{Conditional Video Generation via Flow Matching.}
Flow matching~\cite{esser2024scaling,lipman2022flow,liu2022flow} learns a velocity field that transports a Gaussian distribution to the data distribution.
For video generation, it operates in the latent space of a pretrained variational autoencoder~(VAE), where a video $\mathbf{v}$ is encoded into a latent $\mathbf{z}=\mathcal{E}(\mathbf{v})$ by the encoder $\mathcal{E}$.
Given a clean latent $\mathbf{z}$, Gaussian noise $\boldsymbol{\epsilon}\sim\mathcal{N}(\mathbf{0},\mathbf{I})$, a timestep $t\in[0,1]$, and conditioning signals $\mathbf{y}$~(e.g., text prompts, camera poses, or an initial image), the noisy latent is constructed as $\mathbf{z}^{t}=(1-t)\boldsymbol{\epsilon}+t\mathbf{z}$.
The model is trained to predict the velocity by minimizing:
\begin{equation}
\label{eq:fm_loss}
    \mathcal{L}_{\text{FM}}
    =
    \mathbb{E}_{\mathbf{z},\boldsymbol{\epsilon},\mathbf{y},t}
    \left[
        \left\|
            v_{\theta}(\mathbf{z}^{t}, t, \mathbf{y})
            -
            (\mathbf{z} - \boldsymbol{\epsilon})
        \right\|_2^2
    \right],
\end{equation}
where $v_{\theta}$ denotes the velocity predicted by the model with parameters $\theta$.

\paragraph{Interactive Video Modeling via Teacher Forcing.}
While bidirectional video diffusion models attend to both past and future frames, interactive video models require autoregressive generation, where each video chunk is conditioned on historical context and current user inputs.
Therefore, teacher forcing is used to train this next-chunk prediction process, where each noisy chunk attends only to itself and its clean history. 
Let $\mathcal{Z} = \{\mathbf{z}_1, \mathbf{z}_2, \dots, \mathbf{z}_N\}$ denote the sequence of video latent chunks, where $\mathbf{z}_{i} \in \mathbb{R}^{H \times W \times C}$ is the $i$-th chunk, and let $\mathbf{y}$ denote the user inputs~(e.g., camera poses and text prompts).
Each chunk is autoregressively generated from its history $\mathbf{z}_{<i}$ and the user inputs, which can be formulated as:
\begin{equation}
\label{eq:causal_factorization}
p_\theta(\mathbf{z}_{1:N} \mid \mathbf{y}) = \prod_{i=1}^{N} p_\theta\!\left(\mathbf{z}_i \mid \mathbf{z}_{<i}, \mathbf{y}\right)
\end{equation}
However, teacher forcing departs from the pretraining setting of bidirectional video diffusion models.
In attention, each noisy chunk attends only to itself and its history instead of the full video; in noise level, it is predicted from clean historical latents instead of being jointly denoised with other chunks at a shared timestep.
Both deviations can disrupt pretrained visual priors and degrade generation quality.

\subsection{Mixture-of-Bidirectional-and-Autoregressive Attention}
\label{sec:moba}

\begin{wrapfigure}{r}{0.47\textwidth}
\centering
\includegraphics[width=\linewidth]{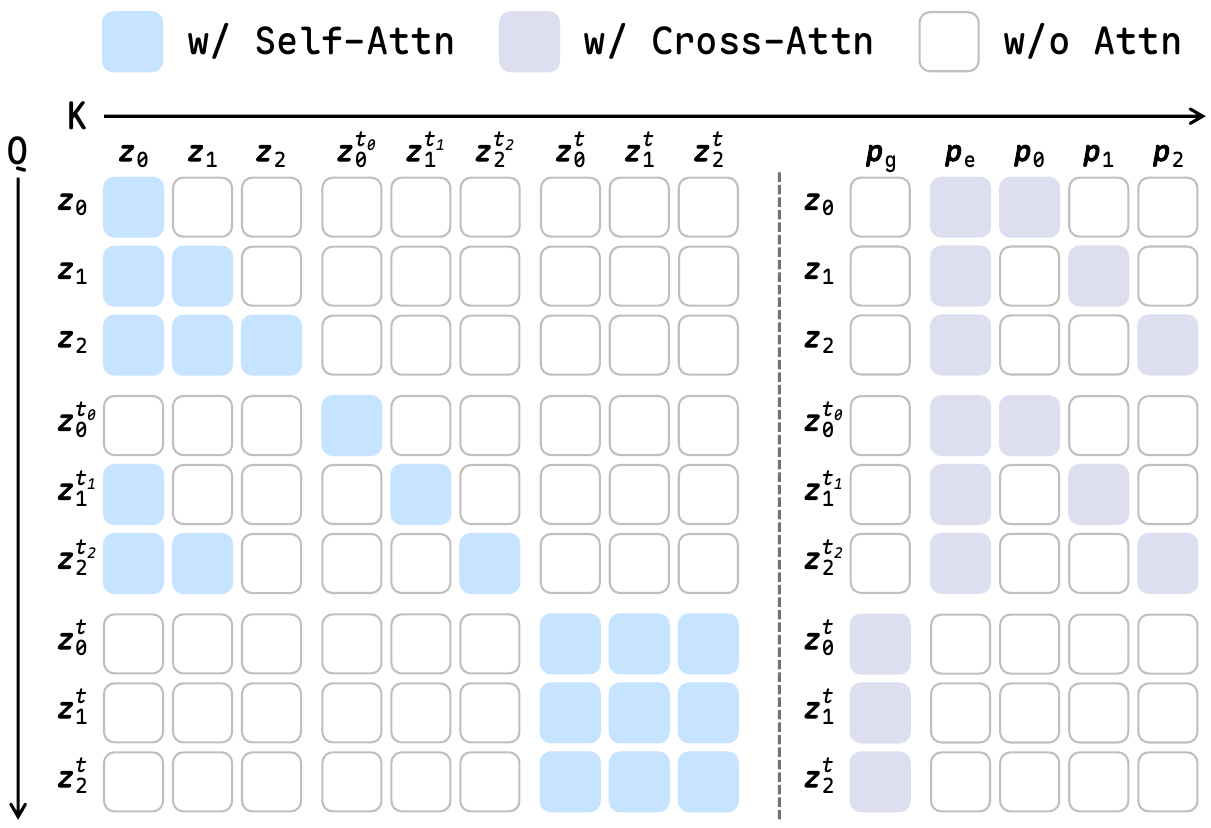}%
\caption{\textbf{\moba{} attention block.} 
For self-attention~(left), each noisy TF chunk attends only to itself and its clean history, while the appended BID noisy latents attend only to one another bidirectionally.
For cross-attention~(right), TF chunks attend to the shared environment prompt $\mathbf{p}_{e}$ and their corresponding local prompt $\mathbf{p}_{i}$, while BID chunks attend to the global prompt $\mathbf{p}_{g}$.
}
\label{fig:moba}
\end{wrapfigure}

As shown in \cref{fig:moba}, the \moba{} attention block is designed for causal pre-training.
Its core is to mix teacher forcing and bidirectional training in self-attention, preserving pretrained visual priors during causal adaptation.
In addition, \moba{} adopts chunk-wise prompt conditioning in cross-attention to enable online text interaction.
We instantiate the user inputs $\mathbf{y}$ in \cref{sec:prelim} as camera poses and text prompts.
Specifically, $\mathbf{e}_i$ denotes the camera pose of the $i$-th chunk, whose representation is detailed in \cref{sec:ivwm}.
The text prompts consist of a shared environment prompt $\mathbf{p}_e$ describing the scene, a local prompt $\mathbf{p}_i$ describing the event in the $i$-th chunk, and a global prompt $\mathbf{p}_g$ describing the entire video.

\paragraph{\moba{} Self-Attention for Preserving Visual Priors.}
To address the deviations discussed in \cref{sec:prelim}, we mix teacher forcing and bidirectional training in self-attention.
Teacher forcing learns causal next-chunk prediction for autoregressive rollout.
Bidirectional training jointly denoises all latents at a shared timestep with full attention, keeping both attention and noise level consistent with pretraining and thus preserving pretrained visual priors.
Formally, for each training sample, we draw a training mode $b\in\{\mathrm{TF},\mathrm{BID}\}$ with equal probability, which defines the following mixture of factorizations~(text conditioning is omitted here and described in the next paragraph):
\begin{equation}
p_\theta^{b}(\mathcal{Z}\mid\mathbf{e}_{1:N})
=
\begin{cases}
\displaystyle\prod_{i=1}^{N}
p_\theta(\mathbf{z}_i\mid
\mathbf{z}_{<i},\mathbf{e}_{\leq i}),
& b=\mathrm{TF}\\[15pt]
p_\theta(\mathbf{z}_{1:N}\mid\mathbf{e}_{1:N}),
& b=\mathrm{BID}
\end{cases}
\label{eq:moba_self_attention}
\end{equation}
In practice, instead of sampling $b$, we append the noisy latents of BID to the TF training sequence and optimize both modes in a single forward pass.
As shown in the left panel of \cref{fig:moba}, a sparse self-attention mask lets each TF noisy chunk attend only to itself and its clean history, while the appended BID latents attend to one another bidirectionally.
Since the two modes are isolated by the mask, this concatenation optimizes the expected objective over $b$ with lower gradient variance.

\paragraph{\moba{} Cross-Attention for Online Text Interaction.}
Online interaction requires the model to update visual content as user instructions arrive.
Existing causal pre-training methods~\cite{zhu2026causal,zhao2026causal} typically condition all chunks on a global prompt describing the entire video, which reveals future events to the current chunk in advance.
Streaming methods such as LongLive~\cite{yang2025longlive} support prompt updates by recaching the recent history under each new prompt, so that past frames are re-encoded with a prompt different from the one they were generated with.
In contrast, we condition each teacher forcing chunk on the environment prompt $\mathbf{p}_e$, which describes the persistent scene, and its own local prompt $\mathbf{p}_i$, which describes the current event, while bidirectional latents remain conditioned on the global prompt $\mathbf{p}_g$, consistent with pretraining.
Formally, the cross-attention output for the $i$-th noisy latent chunk is:
\begin{equation}
\mathbf{o}_i^b =
\begin{cases}
\operatorname{CrossAttn}(\mathbf{z}_i^{t_i},
\mathbf{p}_{e}\oplus\mathbf{p}_{i}), & b=\mathrm{TF}\\[6pt]
\operatorname{CrossAttn}(\mathbf{z}_i^{t},
\mathbf{p}_{g}), & b=\mathrm{BID}
\end{cases}
\label{eq:moba_cross_attention}
\end{equation}
where $\oplus$ denotes concatenation along the token dimension.
In practice, we concatenate the global, shared environment, and all local prompt embeddings into a single sequence.
As shown in the right panel of \cref{fig:moba}, a sparse cross-attention mask lets both clean and noisy latents of chunk $i$ under TF attend only to $\mathbf{p}_{e}$ and $\mathbf{p}_{i}$, while BID latents attend only to $\mathbf{p}_{g}$.
At inference, since each past chunk is encoded with its own local prompt, a prompt update only affects subsequent chunks, and the KV cache of past chunks is reused without recomputation.

\paragraph{\moba{} Training Objective.}
We optimize both modes with the flow matching loss in \cref{eq:fm_loss}.
Under TF, each chunk is noised independently as $\mathbf{z}_i^{t_i}=(1-t_i)\boldsymbol{\epsilon}_i+t_i\mathbf{z}_i$ and predicted from its clean history:
\begin{equation}
\mathcal{L}_{\mathrm{TF}}
=\mathbb{E}\!\left[\frac{1}{N}\sum_{i=1}^{N}
\left\|v_\theta\!\left(\mathbf{z}_i^{t_i},t_i \mid \mathbf{z}_{<i},\mathbf{e}_{\le i},\mathbf{p}_e,\mathbf{p}_i\right)-(\mathbf{z}_i-\boldsymbol{\epsilon}_i)\right\|_2^2\right].
\label{eq:loss_tf}
\end{equation}
In practice, following~\cite{li2026causal}, we perturb the clean history latents with low-level noise with 50\% probability to mitigate error accumulation during autoregressive rollout.
Under BID, all chunks share a timestep $t$ and are jointly denoised:
\begin{equation}
\mathcal{L}_{\mathrm{BID}}
=\mathbb{E}\!\left[\frac{1}{N}
\left\|v_\theta\!\left(\mathbf{z}_{1:N}^{t},t \mid \mathbf{e}_{1:N},\mathbf{p}_g\right)-(\mathbf{z}_{1:N}-\boldsymbol{\epsilon}_{1:N})\right\|_2^2\right].
\label{eq:loss_bid}
\end{equation}
The overall objective is
\begin{equation}
\mathcal{L}_{\mathrm{MoBA}}=\lambda_{\mathrm{TF}}\mathcal{L}_{\mathrm{TF}}+\lambda_{\mathrm{BID}}\mathcal{L}_{\mathrm{BID}},
\label{eq:moba_branch_loss}
\end{equation}
which, with equal weights, is proportional to the expectation over the mode sampling in \cref{eq:moba_self_attention}.

\subsection{\method{} for Interactive Video World Models}
\label{sec:ivwm}
\paragraph{Causal Pre-Training.}
We initialize our interactive video world model from a pretrained bidirectional video model and apply \method{} for causal pre-training.
For camera control, we represent the camera pose $\mathbf{e}_i$ of chunk $i$ with Pl\"ucker embeddings~\cite{he2024cameractrl,he2025cameractrl}, which encode the viewing ray of each pixel as six-dimensional coordinates.
Following~\cite{gao2026infinite,team2026advancing}, we inject them through adaptive layer normalization~(AdaLN)~\cite{xu2019understanding} without disrupting the pretrained visual priors.
During causal pre-training, we jointly train our model with camera-labeled and unlabeled data. For unlabeled data, we use an all-zero dummy {Pl\"ucker embedding} as the condition input.
As a result, a bidirectional video diffusion model is converted into an interactive video model supporting camera control and multi-step autoregressive rollout with KV cache.

\paragraph{Few-Step Post-Training.}
To enable real-time interaction, we further compress our causal pre-trained model into a few-step generator.
Let $\mathbf{z}_{i}^{t_i}$ denote the $i$-th latent chunk at timestep $t_{i}$, and $\mathcal{H}_i=(\mathbf{z}_{<i},\,\mathbf{e}_{\le i},\,\mathbf{p}_e,\,\mathbf{p}_{\le i})$ denote its causal context.
We first use consistency distillation~(CD)~\cite{song2023consistency} to initialize a four-step generator $G_{\theta}$ by minimizing:
\begin{equation}
\label{eq:cd_loss}
\mathcal{L}_{\mathrm{CD}}
=
\mathbb{E}\!\left[
d\!\left(
G_\theta(\mathbf{z}_i^{t_i}, t_{i} \mid \mathcal{H}_i)
,\;
G_{\theta^-}(\tilde{\mathbf{z}}_{i}^{t_i+\Delta t}, t_{i}+\Delta t \mid \mathcal{H}_i)
\right)
\right]
\end{equation}
where $d(\cdot,\cdot)$ is a distance metric, $\tilde{\mathbf{z}}_i^{t_i+\Delta t}$ is obtained by integrating the probability-flow ODE using the causal-pretrained teacher model from time $t_i$ to $t_i+\Delta t$, and the target parameters $\theta^-$ are updated as an exponential moving average~(EMA) of the student parameters $\theta$.

We then refine this four-step generator through distribution matching distillation~(DMD)~\cite{yin2024dmd} to enhance visual fidelity during autoregressive rollout.
Let $\hat{\mathbf{z}}$ denote a video generated under user inputs $\mathcal{U}=(\mathbf{e}_{1:N},\mathbf{p}_e,\mathbf{p}_{1:N})$, and $\hat{\mathbf{z}}^t$ denote its noised version at timestep $t$.
We further optimize $G_{\theta}$ with the following objective:
\begin{equation}
\label{eq:dmd}
\mathcal{L}_{\text{DMD}}
=
\mathbb{E}_{\hat{\mathbf{z}}^t,\hat{\mathbf{z}},\mathcal{U},t}
\left[
\frac{1}{2}
\left\|
\hat{\mathbf{z}}
-
\operatorname{sg}\!\left[
\hat{\mathbf{z}}
-
\left(
s_{\mathrm{fake}}^{\phi}(\hat{\mathbf{z}}^t,t\mid\mathcal{U})
-
s_{\mathrm{real}}(\hat{\mathbf{z}}^t,t\mid\mathcal{U})
\right)
\right]
\right\|_2^2
\right]
\end{equation}
where $s_{\rm real}$ and $s^\phi_{\rm fake}$ represent the real and fake score models, respectively, and $\text{sg}[\cdot]$ denotes the stop-gradient operator.
During DMD training, the parameters $\phi$ of the fake score model are optimized with the standard diffusion loss on generated videos, while the real score model is kept fixed.

\FloatBarrier
\setlength{\parskip}{2.5pt}
\section{Experiments}
\label{sec:experiments}

In this section, we first detail the experimental setup, including benchmarks, evaluation metrics, and implementation details~(\cref{sec:exp_setup}).
We then compare our model with state-of-the-art video world models quantitatively~(\cref{sec:exp_quant}) and qualitatively~(\cref{sec:exp_qual}).
Finally, we conduct an ablation study to validate the design of \moba{} and the effect of \method{} on few-step post-training~(\cref{sec:exp_ab}).

\subsection{Experimental Setup}
\label{sec:exp_setup}
\paragraph{Benchmarks.}
We construct an internal benchmark~(InBench) to evaluate event-following ability and camera-control precision.
We use GPT-Image 2.5~\cite{openai2026gptimage25} to generate 96 initial images and organize InBench into two subsets:
(i) an interactive subset covering environmental changes and character interactions~(32 cases); and
(ii) a roaming subset covering human characters and virtual objects~(64 cases).
Each case is evaluated with a video of 8 to 35 seconds, and each interactive case contains 3 to 5 prompt updates.
We also evaluate on WBench~\cite{ying2026wbench} and SANA-WM-Bench~\cite{zhu2026sanawm}.
\paragraph{Evaluation Metrics.}
Following WBench and SANA-WM-Bench, we report aesthetic quality~(AQ), imaging quality~(IQ), temporal flickering~(TFlk), and dynamic degree~(DD) from VBench~\cite{huang2024vbench}, HPSv3~\cite{ma2025hpsv3}, and rotation error~(Rot, in radians), translation error~(Trans), and camera-motion consistency~(CMC) between the input and estimated camera trajectories.
For long rollouts, $\mathrm{IQ}_{a\text{--}b}$ denotes imaging quality from $a$ to $b$ seconds, and sustained imaging quality~(SIQ) averages the first and last windows.
For online text interaction, we use a vision-language model as a judge~\cite{zheng2023judging} to measure Event Adherence~(EA), the completion of each event, and On-Time Accuracy~(OT), whether a correct response appears within a given time after each prompt update~(see Appendix).
\paragraph{Implementation Details.}
We train on 100k self-collected clips~(974 hours), 70\% of which have camera poses estimated by COLMAP~\cite{schoenberger2016sfm}, while the rest use the dummy Pl\"ucker embedding~(\cref{sec:ivwm}).
For causal pre-training, we initialize from Wan2.1-I2V-14B-480P, train on variable-length $480\times832$ videos at 16 FPS with $\lambda_{\mathrm{TF}}=\lambda_{\mathrm{BID}}=1$, and use AdamW with a learning rate decaying from 5e-5 to 1e-5; the chunk sizes for teacher forcing and chunk-wise cross-attention are 1 and 4 latent frames, respectively.
We randomly perturb the clean context latents using low-level noise with 50\% probability~\cite{li2026causal}.
For post-training, CD uses LPIPS~\cite{zhang2018unreasonable} as the distance metric with a learning rate of 2e-6, and DMD follows LingBot-World-Fast with LingBot-World-Base~\cite{team2026advancing} as score models.
All three stages use 256 H20 GPUs, with 15k, 8k, and 1.5k steps for pre-training, CD, and DMD.

\subsection{Quantitative Comparison}
\label{sec:exp_quant}

\begin{table}[!t]
\centering
\small
\caption{
\textbf{Quantitative comparison on InBench.}
The quality, event-following, and camera-error groups are evaluated on the full InBench, the interactive subset, and the roaming subset, respectively.
Metrics follow \cref{sec:exp_setup}, and Avg denotes the average of the quality metrics.
EA denotes event adherence, and OT\textsubscript{1.0s} and OT\textsubscript{1.5s} denote on-time accuracy within 1.0 and 1.5 seconds.
The best results are in bold, and the top three are highlighted from dark to light blue.
}
\label{tab:inbench}
\begingroup
\fontsize{8}{10}\selectfont
\setlength{\tabcolsep}{0.5pt}
\renewcommand{\arraystretch}{1.15}
\resizebox{0.95\linewidth}{!}{%
\begin{tabularx}{\linewidth}{l*{12}{N}}
\toprule
\multirow{2}{*}{\textbf{Method}} &
\multicolumn{6}{c}{\textbf{Quality} $\boldsymbol{\uparrow}$} &
\multicolumn{3}{c}{\textbf{Event Following} $\boldsymbol{\uparrow}$} &
\multicolumn{3}{c}{\textbf{Camera Error} $\boldsymbol{\downarrow}$} \\
\cmidrule(lr){2-7}\cmidrule(lr){8-10}\cmidrule(l){11-13}
& \textbf{AQ} & \textbf{IQ} & \textbf{TFlk} &
\textbf{DD} & \textbf{HPSv3} &
\textbf{Avg} & \textbf{EA} &
\textbf{OT}\textsubscript{1.0s} &
\textbf{OT}\textsubscript{1.5s} &
\textbf{Rot} & \textbf{Trans} & \textbf{CMC} \\
\midrule
HY-WorldPlay
& 59.4 & 62.4 & 96.8 & 55.6 & 58.4 & 66.5
& 0.0 & 0.0 & 0.0
& 1.795 & 0.553 & 1.968 \\
JoyAI-Echo-WM
& 62.7 & 62.5 & 95.4 & 74.6 & 62.2 & 71.5
& \cellcolor{tableRank3}45.2
& \cellcolor{tableRank3}5.6
& \cellcolor{tableRank3}26.2
& \cellcolor{tableRank3}0.928
& 0.463
& \cellcolor{tableRank3}1.126 \\
Matrix-Game 3.0
& 60.6 & 65.2 & 95.3 & 87.3 & 60.7
& \cellcolor{tableRank2}73.8
& 0.0 & 0.0 & 0.0
& \cellcolor{tableRank2}0.441
& \cellcolor{tableRank2}0.329
& \cellcolor{tableRank2}0.694 \\
SANA-WM-Streaming
& 61.0 & 56.9 & 96.6 & 79.4 & 59.1 & 70.6
& 41.3 & 4.8 & 23.0
& 2.022 & 0.466 & 2.067 \\
LingBot-World-Fast
& 63.8 & 61.7 & 96.3 & 73.0 & 67.4
& \cellcolor{tableRank3}72.4
& \cellcolor{tableRank2}86.5
& \cellcolor{tableRank2}38.1
& \cellcolor{tableRank2}87.3
& 1.065
& \cellcolor{tableRank3}0.433
& 1.312 \\
\textbf{\method{}~(Ours)}
& 63.2 & 63.8 & 96.0 & 81.0 & 71.3
& \cellcolor{tableRank1}\textbf{75.1}
& \cellcolor{tableRank1}\textbf{94.4}
& \cellcolor{tableRank1}\textbf{88.9}
& \cellcolor{tableRank1}\textbf{96.8}
& \cellcolor{tableRank1}\textbf{0.296}
& \cellcolor{tableRank1}\textbf{0.316}
& \cellcolor{tableRank1}\textbf{0.539} \\
\bottomrule
\end{tabularx}%
}%
\endgroup
\end{table}

\paragraph{Evaluation on InBench.}
\Cref{tab:inbench} shows that our model achieves the best average quality, event-following, and camera-control results among all methods.
The highest average quality score of 75.1 indicates that \method{} preserves pretrained visual priors during causal pre-training, and the event-following scores~(EA=94.4, OT\textsubscript{1.5s}=96.8) validate that chunk-wise cross-attention enables effective responses to online prompts.
The lowest camera errors further demonstrate precise camera control during autoregressive rollout.

\begin{table}[!t]
\centering
\caption{\textbf{Quantitative comparison on SANA-WM-Bench.}
60-second generation under (a) simple and (b) hard camera trajectories.
$\mathrm{IQ}_{0\text{--}10}$ and $\mathrm{IQ}_{50\text{--}60}$ denote imaging quality over the first and last 10 seconds.
Other metrics follow \cref{sec:exp_setup}.
The best results are in bold, and the top three are highlighted from dark to light blue.
}
\label{tab:sanawm}
\begingroup
\fontsize{8}{9.5}\selectfont
\setlength{\tabcolsep}{1.25pt}
\renewcommand{\arraystretch}{1.15}
\captionsetup[subtable]{position=top,skip=3pt}
\begin{subtable}[t]{0.49\linewidth}
\centering
\caption{Simple trajectory splits.}
\label{tab:sanawm_simple}
\resizebox{\linewidth}{!}{%
\begin{tabularx}{275pt}{@{}l*{7}{N}@{}}
\toprule
\multirow{2}{*}{\textbf{Method}} & \multirow{2}{*}{\shortstack[c]{\textbf{Quality} $\boldsymbol{\uparrow}$}} & \multicolumn{3}{c}{\textbf{Temporal IQ} $\boldsymbol{\uparrow}$} & \multicolumn{3}{c}{\textbf{Camera Error} $\boldsymbol{\downarrow}$} \\
\cmidrule(lr){3-5}\cmidrule(l){6-8}
& & $\mathrm{IQ}_{0\text{--}10}$ & $\mathrm{IQ}_{50\text{--}60}$ & \textbf{SIQ} & \textbf{Rot} & \textbf{Trans} & \textbf{CMC} \\
\midrule
WorldPlay & 61.7 & 68.7 & 61.6 & 65.0 & 1.087 & \rankthird{1.552} & 2.240 \\
Echo-WM & 69.4 & 67.6 & 64.4 & \ranksecond{66.0} & \ranksecond{0.902} & \rankfirst{1.450} & \rankfirst{2.010} \\
Matrix-Game 3 & \ranksecond{70.4} & 65.3 & 65.5 & \rankthird{65.4} & \rankfirst{0.888} & 1.575 & \rankthird{2.118} \\
SANA-Streaming & 69.7 & 65.2 & 59.4 & 62.2 & 1.374 & 1.768 & 2.558 \\
LingBot-Fast & \rankthird{70.1} & 63.2 & 52.0 & 57.1 & 1.964 & 1.616 & 2.836 \\
\textbf{\method{}~(Ours)} & \rankfirst{74.8} & 68.6 & 64.9 & \rankfirst{66.7} & \rankthird{0.965} & \ranksecond{1.459} & \ranksecond{2.038} \\
\bottomrule
\end{tabularx}%
}%
\end{subtable}\hfill
\begin{subtable}[t]{0.49\linewidth}
\centering
\caption{Hard trajectory splits.}
\label{tab:sanawm_hard}
\resizebox{\linewidth}{!}{%
\begin{tabularx}{275pt}{@{}l*{7}{N}@{}}
\toprule
\multirow{2}{*}{\textbf{Method}} & \multirow{2}{*}{\shortstack[c]{\textbf{Quality} $\boldsymbol{\uparrow}$}} & \multicolumn{3}{c}{\textbf{Temporal IQ} $\boldsymbol{\uparrow}$} & \multicolumn{3}{c}{\textbf{Camera Error} $\boldsymbol{\downarrow}$} \\
\cmidrule(lr){3-5}\cmidrule(l){6-8}
& & $\mathrm{IQ}_{0\text{--}10}$ & $\mathrm{IQ}_{50\text{--}60}$ & \textbf{SIQ} & \textbf{Rot} & \textbf{Trans} & \textbf{CMC} \\
\midrule
WorldPlay & 65.5 & 68.7 & 63.5 & \ranksecond{66.0} & \rankthird{1.500} & \rankthird{1.550} & \rankthird{2.525} \\
Echo-WM & \ranksecond{72.9} & 67.4 & 64.4 & \rankthird{65.9} & 1.529 & 1.589 & 2.597 \\
Matrix-Game 3 & 70.2 & 65.2 & 66.7 & \rankthird{65.9} & \rankfirst{0.914} & \ranksecond{1.520} & \rankfirst{2.101} \\
SANA-Streaming & \rankthird{71.1} & 64.6 & 59.9 & 62.2 & 1.637 & 1.647 & 2.687 \\
LingBot-Fast & 69.9 & 63.4 & 54.3 & 58.5 & 1.871 & 1.683 & 2.887 \\
\textbf{\method{}~(Ours)} & \rankfirst{74.7} & 68.5 & 64.5 & \rankfirst{66.4} & \ranksecond{1.214} & \rankfirst{1.456} & \ranksecond{2.246} \\
\bottomrule
\end{tabularx}%
}%
\end{subtable}
\endgroup
\end{table}

\paragraph{Evaluation on Public Benchmarks.}
We report SANA-WM-Bench results in \cref{tab:sanawm}.
On both the simple and hard 60-second splits, our model achieves the highest average \textit{Quality}~(74.8 and 74.7) and SIQ~(66.7 and 66.4) scores, whereas several baselines show clear quality drops in later windows, e.g., the IQ of LingBot-World-Fast decreases from 63.2 to 52.0 on the simple split.
These results demonstrate that our model maintains visual fidelity throughout minute-long autoregressive generation; one-hour qualitative results and average quality scores measured every five minutes are further provided in Figure.~S12 in the Appendix.
Although competitive, our camera errors are not the lowest on SANA-WM-Bench.
We partly attribute this gap to a prompt mismatch: the benchmark describes the entire video with a single global prompt, whereas our model is trained with an environment prompt and chunk-wise local prompts.
Although local prompts do not explicitly specify camera motion, they indicate what should appear in the current chunk and thus help the pose condition control viewpoints more precisely, as supported by the reduced camera errors with chunk-wise prompts in \cref{tab:cross_attn_ablation}.
We further evaluate on the Navi-158 subset of WBench~(Table.~S1 in the Appendix), where our model achieves the highest average \textit{Quality}~(78.5) and \textit{Setting} adherence~(80.3) scores, indicating strong visual fidelity and faithful reflection of the specified world settings, while its \textit{Interaction} scores show a gap similar to that of the camera errors on SANA-WM-Bench.

\paragraph{User Study.}
To complement these automatic metrics, we further conduct a pairwise user study.
Participants compare our model with one baseline at a time in a method-blinded, side-by-side setting, rating visual quality on all cases, interactive following on the interactive subset, and motion clarity on the roaming subset.
We collect 318 pairwise trials from 18 sessions, with details provided in the Appendix.
As shown in \cref{fig:user_study}, participants consistently prefer our model over each baseline in terms of visual quality, interactive following, and motion clarity.

\begin{figure}[t!]
\centering
\includegraphics[width=0.95\textwidth]{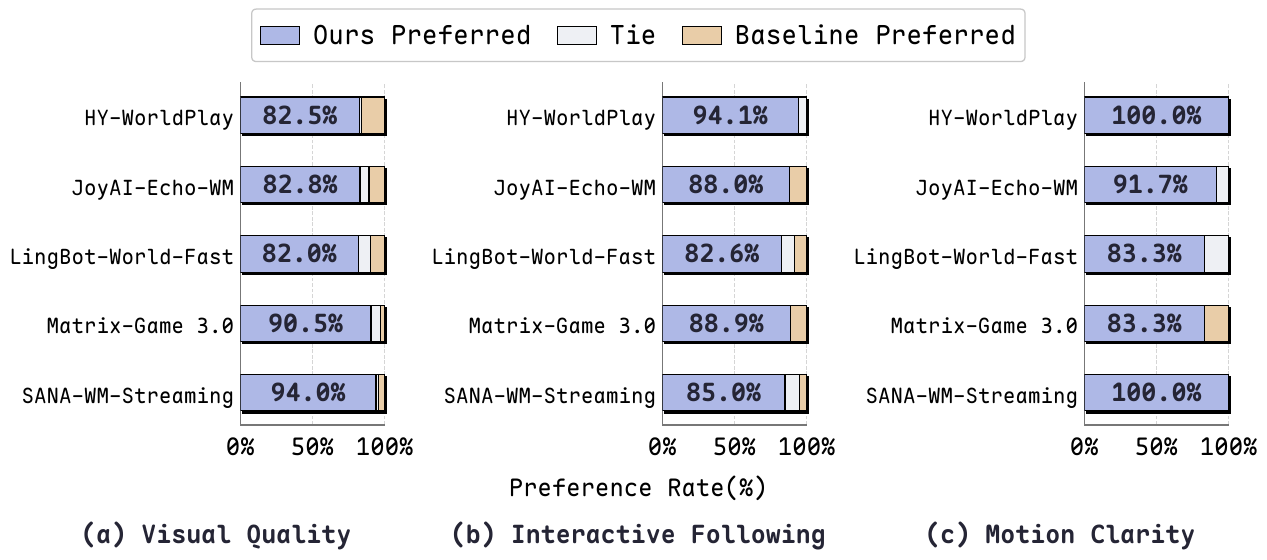}%
\caption{\textbf{Pairwise user study on InBench.}
Each bar compares our model with one baseline and shows the percentages of judgments preferring ours, rating a tie, and preferring the baseline.
}
\label{fig:user_study}
\end{figure}

\begin{figure}[t!]
\centering
\captionsetup[subfigure]{skip=1pt}
\begin{subfigure}{0.98\textwidth}
\centering
\includegraphics[width=\linewidth]{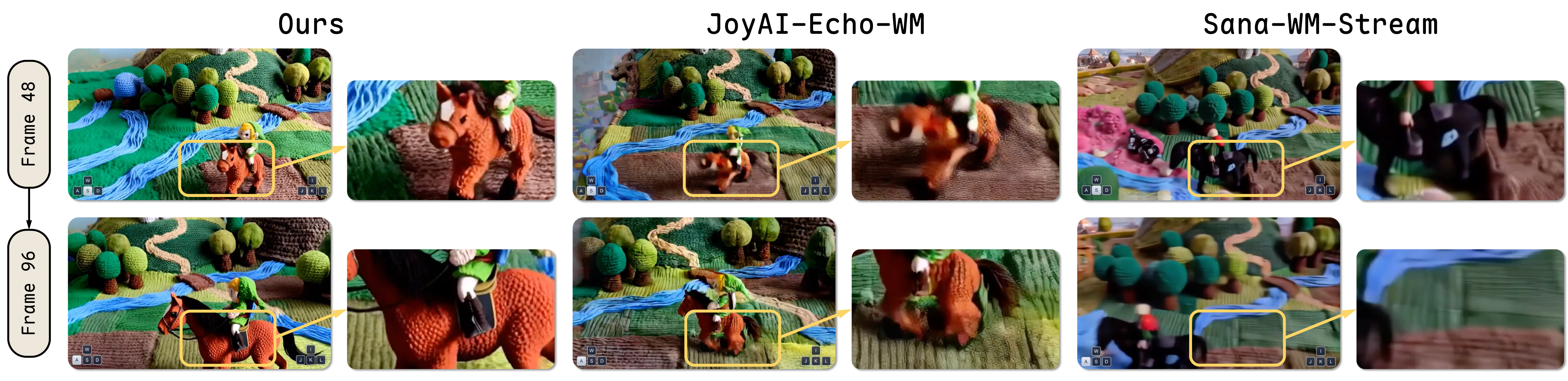}%
\caption{Camera-controlled roaming.}
\label{fig:inbench_roam_qualitative}
\end{subfigure}
\par\vspace{5pt}
\begin{subfigure}{0.98\textwidth}
\centering
\includegraphics[width=\linewidth]{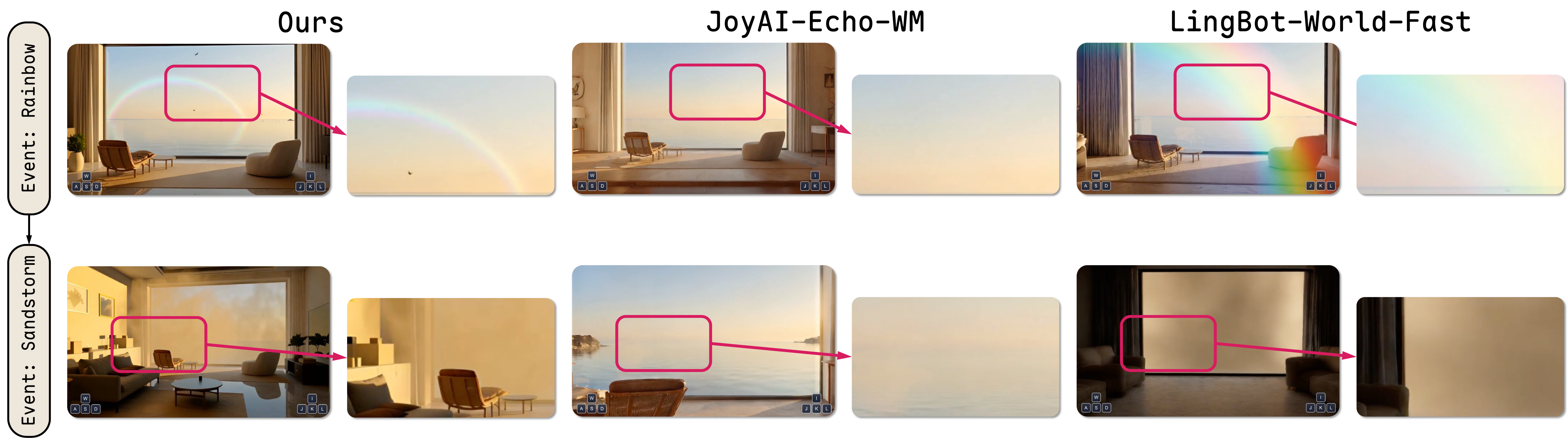}%
\caption{Online prompt interaction.}
\label{fig:inbench_inter_qualitative}
\end{subfigure}
\caption{\textbf{Qualitative comparisons with state-of-the-art methods.}
(a) Camera-controlled roaming: our model preserves structure and fine details during viewpoint changes, whereas other methods suffer from blurring or distortion.
(b) Online prompt interaction: our model follows each event prompt with a consistent scene, whereas other methods respond with delays or break scene consistency.
}
\label{fig:inbench_qualitative}
\end{figure}

\subsection{Qualitative Comparison}
\label{sec:exp_qual}

\paragraph{Camera-Controlled Roaming.}
Fig.~\ref{fig:inbench_qualitative}\subref{fig:inbench_roam_qualitative} illustrates the qualitative comparison on the InBench roaming subset.
During camera movement, especially when the viewpoint changes, the video model needs to generate the visual content from a novel perspective while keeping its structure and appearance stable across frames.
We observe that existing state-of-the-art methods suffer from progressive blurring and structural distortion, recovering sharpness only after the camera stops moving.
In contrast, the enlarged regions highlighted by yellow boxes show that our model better maintains scene consistency, motion clarity, and fine-grained appearance throughout the viewpoint transition.

\paragraph{Online Prompt Interaction.}
Fig.~\ref{fig:inbench_qualitative}\subref{fig:inbench_inter_qualitative} illustrates the qualitative comparison on the InBench interactive subset.
Although JoyAI-Echo-WM and LingBot-World-Fast support prompt updates, we observe that they respond with noticeable delays, consistent with their lower OT\textsubscript{1.0s} in \cref{tab:inbench}, or disrupt existing scene content.
In contrast, with the chunk-wise cross-attention in \method{}, the enlarged regions highlighted by red boxes show that our model responds to each local prompt and generates the correct content while maintaining scene consistency.

\subsection{Ablation Study}
\label{sec:exp_ab}
For the ablation study, we sample a subset of our internal training data with each video clip containing 161 frames and follow the training settings described in \cref{sec:exp_setup}.
The evaluation is carried out on InBench, where we measure generation quality on the full benchmark, event-following ability on the interactive subset, and camera error on the roaming subset.

\begin{table}[!htbp]
\centering
\small
\caption{\textbf{Ablation of \moba{} self-attention design.}
Combinations of teacher forcing~(TF), diffusion forcing~(DF), and bidirectional~(BID) training with different training chunk sizes.
$\mathrm{IQ}_{0\text{--}2}$ and $\mathrm{IQ}_{8\text{--}10}$ denote imaging quality over the first and last 2 seconds.
The best results are in bold, and the top three are highlighted from dark to light blue.
}
\label{tab:mask_ablation}
\begingroup
\small
\setlength{\tabcolsep}{2pt}
\renewcommand{\arraystretch}{1.15}
\resizebox{0.95\linewidth}{!}{%
\begin{tabularx}{\linewidth}{lc*{7}{N}}
\toprule
\multicolumn{1}{c}{\multirow{2}{*}{\textbf{Attention Pattern}}} &
\multirow{2}{*}{\textbf{Chunk}} &
\multirow{2}{*}{\textbf{Quality} $\boldsymbol{\uparrow}$} &
\multicolumn{3}{c}{\textbf{Temporal IQ} $\boldsymbol{\uparrow}$} &
\multicolumn{3}{c}{\textbf{Camera Error} $\boldsymbol{\downarrow}$} \\
\cmidrule(lr){4-6}\cmidrule(l){7-9}
& & &
$\boldsymbol{\mathrm{IQ}_{0\text{--}2}}$ &
$\boldsymbol{\mathrm{IQ}_{8\text{--}10}}$ &
$\boldsymbol{\mathrm{SIQ}}$ &
\textbf{Rot} & \textbf{Trans} & \textbf{CMC} \\
\midrule
TF                    & 1   & 68.9 & 51.5 & 48.3 & 49.9 & 1.098 & 0.445 & 1.425 \\
TF+DF                 & 1   & 65.6 & 52.8 & 50.9 & 51.9 & \cellcolor{tableRank2}0.655 & \cellcolor{tableRank2}0.249 & \cellcolor{tableRank2}0.871 \\
TF+DF+BID             & 1   & 67.7 & 52.1 & 48.8 & 50.5 & \cellcolor{tableRank3}0.751 & \cellcolor{tableRank3}0.258 & \cellcolor{tableRank3}0.986 \\
TF+BID                & 1   & \cellcolor{tableRank1}\textbf{70.8} & 55.9 & 61.0 & \cellcolor{tableRank1}\textbf{58.4} & \cellcolor{tableRank1}\textbf{0.449} & \cellcolor{tableRank1}\textbf{0.231} & \cellcolor{tableRank1}\textbf{0.647} \\
TF+BID                & 4   & \cellcolor{tableRank2}69.1 & 55.2 & 58.8 & \cellcolor{tableRank2}57.0 & 0.884 & 0.323 & 1.165 \\
TF+BID                & 1+4 & \cellcolor{tableRank3}69.1 & 54.0 & 55.9 & \cellcolor{tableRank3}54.9 & 0.882 & 0.331 & 1.162 \\
\bottomrule
\end{tabularx}%
}%
\endgroup
\end{table}

\Needspace{5\baselineskip}
\paragraph{Self-Attention Design for Causal Pre-Training.} 
Using teacher forcing~(TF) as the baseline, we evaluate the effects of adding diffusion forcing~(DF)~\cite{chen2024diffusion} and bidirectional~(BID) training.
As shown in \cref{tab:mask_ablation}, TF+BID~(row 4) improves generation quality, achieves the highest SIQ of 58.4, and maintains high final-window quality (from IQ\textsubscript{0--2}=55.9 to IQ\textsubscript{8--10}=61.0), while IQ drops from 51.5 to 48.3 under TF, indicating that bidirectional attention can mitigate quality degradation.
However, adding DF~(rows 2 and 3) can reduce generation quality.
We attribute this degradation to a mismatch between training and inference, where DF uses noisy historical chunks during training, while KV-cached inference leverages fully denoised historical context.
Note that DF replaces the clean history with independently noised history, which matches neither the clean KV cache at inference nor the shared noise level of pretraining. In contrast, \moba{} keeps the clean history in TF for inference consistency and recovers the pretraining setting in a separate BID branch.
We further study the effect of training chunk size under TF+BID.
As chunk size=1~(row 4) enables fine-grained causal modeling by assigning an independent noise level to each latent frame and outperforms the other settings~(rows 5 and 6) in both generation quality and camera control accuracy, we combine TF and BID training with chunk size=1 in \moba{}.

\begin{table}[!htbp]
\centering
\small
\caption{\textbf{Ablation of \moba{} cross-attention design.}
Global versus chunk-wise prompt conditioning, both built on the best self-attention setting in \cref{tab:mask_ablation}.
Metrics follow \cref{sec:exp_setup}.
The best results are in bold and highlighted in blue.
}
\label{tab:cross_attn_ablation}
\begingroup
\small
\setlength{\tabcolsep}{2pt}
\renewcommand{\arraystretch}{1.15}
\resizebox{0.95\linewidth}{!}{%
\begin{tabularx}{\linewidth}{l*{6}{N}}
\toprule
\multirow{2}{*}{\textbf{Prompt}} &
\multirow{2}{*}{\textbf{Quality} $\boldsymbol{\uparrow}$} &
\multicolumn{2}{c}{\textbf{Event Following} $\boldsymbol{\uparrow}$} &
\multicolumn{3}{c}{\textbf{Camera Error} $\boldsymbol{\downarrow}$} \\
\cmidrule(lr){3-4}\cmidrule(l){5-7}
& &
\textbf{EA} & \textbf{OT}\textsubscript{1.5s} &
\textbf{Rot} & \textbf{Trans} & \textbf{CMC} \\
\midrule
Global
& 70.8
& 23.4 & 18.3
& 0.449 & 0.231 & 0.647 \\
Chunk-Wise
& \cellcolor{tableRank1}\textbf{71.1}
&  \cellcolor{tableRank1}\textbf{50.0} &  \cellcolor{tableRank1}\textbf{62.1}
& \cellcolor{tableRank1}\textbf{0.319}
& \cellcolor{tableRank1}\textbf{0.139}
& \cellcolor{tableRank1}\textbf{0.431} \\
\bottomrule
\end{tabularx}%
}%
\endgroup
\end{table}

\paragraph{Cross-Attention Design for Causal Pre-Training.} 
As shown in \cref{tab:cross_attn_ablation}, we study the effect of global and chunk-wise prompts in cross-attention.
Chunk-wise prompts improve event-following ability, increasing EA from 23.4 to 50.0 and OT\textsubscript{1.5s} from 18.3 to 62.1.
Chunk-wise prompts also reduce all three camera errors, consistent with our analysis in \cref{sec:exp_quant}.

\begin{table}[!htbp]
\centering
\small
\caption{\textbf{Ablation of \method{} for few-step post-training.}
Four-step models obtained from TF-only and \method{} pre-training under the same CD and DMD pipeline, with the 50-step \method{} model as a reference.
Metrics follow \cref{sec:exp_setup}.
The best results are in bold and highlighted in blue.
}
\label{tab:post_ablation}
\begingroup
\fontsize{8}{10}\selectfont
\setlength{\tabcolsep}{2pt}
\renewcommand{\arraystretch}{1.15}
\resizebox{0.95\linewidth}{!}{%
\begin{tabularx}{\linewidth}{ccc*{6}{N}}
\toprule
\multirow{2}{*}[-2.7pt]{\textbf{Pre-training}} &
\multirow{2}{*}[-2.7pt]{\textbf{Post-training}} &
\multirow{2}{*}[-2.7pt]{\textbf{Steps}} &
\multirow{2}{*}[-2.7pt]{\textbf{Quality} {\boldmath$\uparrow\kern-\fontcharic\textfont2"22$}} &
\multicolumn{2}{c}{\textbf{Event Following} $\boldsymbol{\uparrow}$} &
\multicolumn{3}{c}{\textbf{Camera Error} $\boldsymbol{\downarrow}$} \\
\cmidrule(lr){5-6}\cmidrule(l){7-9}
& & & &
\textbf{EA} &
\textbf{OT}\textsubscript{1.5s} &
\textbf{Rot} & \textbf{Trans} & \textbf{CMC} \\
\midrule
\method{} & None                 & 50 & \cellcolor{tableRank1}\textbf{71.1} & 50.0 & 62.1 & \cellcolor{tableRank1}\textbf{0.319} & \cellcolor{tableRank1}\textbf{0.139} & \cellcolor{tableRank1}\textbf{0.431} \\
TF only   & CD $\rightarrow$ DMD &  4 & 64.1 & 33.7 & 19.8 & 1.077 & 0.446 & 1.331 \\
\method{} & CD $\rightarrow$ DMD &  4 & 70.9 & \cellcolor{tableRank1}\textbf{55.1} & \cellcolor{tableRank1}\textbf{64.3} & 0.435 & 0.234 & 0.593 \\
\bottomrule
\end{tabularx}%
}%
\endgroup
\end{table}

\paragraph{\method{} for Few-Step Post-Training.} 
As shown in \cref{tab:post_ablation}, we study the effect of pre-training with \method{} on few-step post-training.
Under the same CD and DMD pipeline, the four-step model pretrained with \method{}~(row 3) outperforms the TF-only baseline~(row 2), indicating that \method{} provides a stronger initialization.
Compared with the 50-step model~(row 1), the four-step model retains comparable generation quality and achieves higher event-following scores.

\FloatBarrier
\Needspace{10\baselineskip}

\section{Conclusion}
\label{sec:conclusion}
We present \method{}, a causal pre-training paradigm for interactive video world models.
Implemented with the \moba{} attention block, it mixes teacher forcing and bidirectional training to preserve pretrained visual priors, and uses chunk-wise prompts for online text interaction.
Combined with CD and DMD, our four-step model achieves state-of-the-art visual quality and timely event following with camera control over one-hour rollouts.

\paragraph{Limitations.}
Our implementation targets video DiT architectures with separate self-attention and cross-attention modules, and its extension to unified attention~\cite{liang2024mixture} remains to be verified.
Moreover, our model is designed for chunk-wise local prompts, and its camera control is less precise when only a single global prompt is available, as observed on public benchmarks.
We leave these directions for future work.

\ifdefined\BuildCombinedPaper
  \clearpage
  \appendix
  \mixforcingsupplementarylayout
  \section*{Supplementary Material}
  \noindent Visual comparison results are available at \url{https://zelingao98.github.io/mixforcing}.
  \supplementarycontents
  
\section{Implementation Details}
\label{app:implementation}

\subsection{Training Data}
\label{app:training_data}
\paragraph{Data Acquisition.}
We collect 100k video clips totaling 974 hours.
We collect egocentric recordings, videos from games and Unreal Engine environments, and web videos.
These sources cover first-person interactions, controlled synthetic scenes, and diverse real-world content.
\paragraph{Data Filtering.}
We split videos at shot boundaries and remove clips with decoding errors or unsuitable duration and resolution.
We further filter clips using visual quality, exposure, text occupancy, and motion statistics.
A vision-language model assesses the remaining clips for editing artifacts, viewpoint, and interaction content.
\paragraph{Prompt Annotation.}
For each clip, we generate a shared environment prompt describing the scene and a global prompt summarizing the entire video.
We separately annotate subject visibility, motion, interactions, and environmental changes, then combine the active annotations into local prompts.
We refine these prompts for consistent wording and temporal alignment, removing descriptions of future events or actions unsupported by visual evidence.
Each local prompt is shared by four consecutive latent frames, consistent with the cross-attention setting in the main paper.
\paragraph{Camera Annotation.}
We estimate camera poses using COLMAP for 70\% of the video clips.
The remaining 30\% are used without camera annotations, with all-zero Pl\"ucker embeddings as the conditioning input.

\subsection{Training Settings}
\label{app:training_settings}
\paragraph{Causal Pre-Training.}
We initialize our model from Wan2.1-I2V-14B-480P and train on variable-length videos at 16 FPS with a resolution of $480\times832$.
We train for 15k steps on 512 H200 GPUs using AdamW with a learning rate of $5\times10^{-5}$ decaying to $1\times10^{-5}$.
TF and BID are jointly optimized in a single forward pass with equal loss weights, $\lambda_{\mathrm{TF}}=\lambda_{\mathrm{BID}}=1.0$.
Under TF, we use chunk size=1 and independently sample a timestep for each latent frame, while BID uses a shared timestep for the entire video.
We set the flow shift to 5.
We perturb the clean historical context with low-level noise with 50\% probability to improve robustness during autoregressive rollout~\cite{li2026causal}.
\paragraph{Consistency Distillation.}
We use the causal pre-trained model as the teacher to distill a four-step generator.
We fully fine-tune the video backbone and camera-control modules, using chunk size=4 latent frames and conditioning on one initial frame.
We optimize the student for 8k steps on 512 H20 GPUs using AdamW with a learning rate of $2\times10^{-6}$, $(\beta_1,\beta_2)=(0,0.999)$, and no learning-rate warm-up.
The target model is maintained as an exponential moving average of the student parameters with a decay rate of 0.99.
We set $N=48$ for discrete CD and use a guidance scale of 3.0.
We replace use LPIPS as CD loss.
\paragraph{Distribution Matching Distillation.}
We further refine the four-step generator obtained from CD using DMD for 1.5k steps on 512 H20 GPUs.
We retain the same four-step schedule with timestep indices $\{0,250,500,750\}$.
We use LingBot-World-Base~\cite{team2026advancing} for the real and fake score models.
The real score model remains fixed, while the fake score model is updated with the standard diffusion loss on generated videos.
The generator is optimized using the distribution matching objective described in the main paper.
We set the learning rates to $2\times10^{-6}$ for the generator and $4\times10^{-7}$ for the fake score model, with $(\beta_1,\beta_2)=(0,0.999)$ for both optimizers.
The fake score model is updated at every training step, while the generator is updated once every five steps.

\section{Evaluation Details}
\label{app:evaluation}

\subsection{Quality and Camera-Control Metrics}
\label{app:quality_camera_metrics}
We adopt the quality metrics from VBench~\cite{huang2024vbench} and WBench~\cite{ying2026wbench}, and the camera-error metrics from SANA-WM-Bench~\cite{zhu2026sanawm}.
On InBench, quality is evaluated on the full benchmark, while camera errors are evaluated on the roaming subset.
Quality scores are reported on a 0--100 scale, with higher values indicating better performance; lower camera errors indicate more accurate control.

\paragraph{Aesthetic Quality (AQ).}
AQ evaluates the visual appeal of generated frames, including composition, color, and lighting.
The LAION aesthetic predictor scores the normalized CLIP ViT-L/14 features of each sampled frame.
We average these frame scores within each video and multiply the result by 10 to convert the original 0--10 scale to 0--100.

\paragraph{Imaging Quality (IQ).}
IQ evaluates frame quality, including sharpness, noise, and visual artifacts.
We use MUSIQ, a no-reference image-quality model, to score each sampled frame after resizing its longer edge to at most 512 pixels.
The video score is the mean of the frame scores on the 0--100 scale.

\paragraph{Temporal Flickering (TFlk).}
TFlk measures frame-to-frame appearance variation using the mean absolute pixel difference between consecutive evaluated frames.
Let $\bar{d}$ denote this difference averaged over frame pairs, spatial locations, and color channels, with pixel values in $[0,255]$.
The score is $\mathrm{TFlk}=100(1-\bar{d}/255)$, so a higher score indicates less flickering.

\paragraph{Dynamic Degree (DD).}
DD measures the proportion of generated videos containing substantial motion.
We estimate optical flow between consecutive sampled frames using RAFT and average the largest 5\% of flow magnitudes for each pair.
Following the official implementation, the motion threshold is $6\min(H,W)/256$, where $H$ and $W$ are the frame dimensions after preprocessing.
A video with $N$ sampled frames is classified as dynamic if at least $\max(1,\operatorname{round}(N/4))$ frame pairs exceed this threshold.
DD is the percentage of videos classified as dynamic.

\paragraph{Human Preference Score v3 (HPSv3).}
HPSv3 estimates human visual preference using a learned reward model.
Following WBench, we evaluate up to 20 uniformly sampled frames per video with empty text prompts and average their raw rewards to obtain $\bar{r}$.
We use the fixed normalization bounds from the official implementation:
\begin{equation}
\mathrm{HPSv3}
=100\operatorname{clip}\!\left(\frac{\bar{r}-r_{\mathrm{low}}}{r_{\mathrm{high}}-r_{\mathrm{low}}},0,1\right),
\quad r_{\mathrm{low}}=-5.21,\quad r_{\mathrm{high}}=8.61.
\label{eq:supp_hpsv3}
\end{equation}
These bounds are shared across models and are not re-estimated for each model.

\paragraph{Quality Average.}
We average the per-video scores of each metric over the evaluated videos.
The \textit{Quality} score reported in the comparison and ablation tables is the arithmetic mean of the five metrics:
\begin{equation}
\mathrm{Quality}
=\frac{\mathrm{AQ}+\mathrm{IQ}+\mathrm{TFlk}+\mathrm{DD}+\mathrm{HPSv3}}{5}.
\label{eq:supp_quality_average}
\end{equation}

\paragraph{Temporal IQ and Sustained Imaging Quality (SIQ).}
Temporal IQ evaluates how imaging quality changes during autoregressive rollout by applying the same IQ metric to different time windows.
For the 10-second ablation videos, we evaluate the first and last two seconds, denoted by $\mathrm{IQ}_{0\text{--}2}$ and $\mathrm{IQ}_{8\text{--}10}$.
For the 60-second SANA-WM-Bench videos, we use $\mathrm{IQ}_{0\text{--}10}$ and $\mathrm{IQ}_{50\text{--}60}$.
In both settings, SIQ is the arithmetic mean of the first- and last-window IQ scores.
We report both window scores alongside SIQ to show the change in quality over time.

\newpage
\paragraph{Camera Pose Estimation.}
Following SANA-WM-Bench, we estimate camera poses from generated videos using Pi3X and compare them with the input camera trajectories at corresponding frames.
The trajectories are expressed relative to their first frame, and the estimated trajectory is aligned to the target trajectory in rotation, translation, and scale using the official evaluation procedure.
Let $(\mathbf{R}_k,\mathbf{t}_k)$ and $(\hat{\mathbf{R}}_k,\hat{\mathbf{t}}_k)$ denote the target and aligned estimated camera-to-world poses, respectively, for evaluated frame $k$.
We compute the following errors over $M$ evaluated frames and average the video-level results across videos.

\paragraph{Rotation Error (Rot).}
Rot measures the angular difference between the estimated and target camera orientations.
We report the mean geodesic rotation error in radians, consistent with the main-paper tables:
\begin{equation}
\mathrm{Rot}
=\frac{1}{M}\sum_{k=1}^{M}
\arccos\!\left[\operatorname{clip}\!\left(
\frac{\operatorname{tr}(\mathbf{R}_k^{\top}\hat{\mathbf{R}}_k)-1}{2},-1,1
\right)\right].
\label{eq:supp_rotation_error}
\end{equation}

\paragraph{Translation Error (Trans).}
Trans measures the positional difference between the aligned estimated and target camera centers.
It is computed as $\mathrm{Trans}=\frac{1}{M}\sum_{k=1}^{M}\|\hat{\mathbf{t}}_k-\mathbf{t}_k\|_2$, in the coordinate scale of the target trajectory.

\paragraph{Camera-Motion Consistency (CMC).}
CMC jointly measures rotation and translation errors using the difference between the full pose matrices.
With $\mathbf{P}_k=[\mathbf{R}_k\mid\mathbf{t}_k]$ and $\hat{\mathbf{P}}_k=[\hat{\mathbf{R}}_k\mid\hat{\mathbf{t}}_k]$, we compute $\mathrm{CMC}=\frac{1}{M}\sum_{k=1}^{M}\|\hat{\mathbf{P}}_k-\mathbf{P}_k\|_F$, where $\|\cdot\|_F$ denotes the Frobenius norm.

\subsection{Event-Following Evaluation}
\label{app:event_eval}
We evaluate event-following ability on the interactive subset of InBench using a vision-language model.
For each requested event, the evaluator records its completion score and the timestamp of the first visible response.
EA measures event completion, while $\mathrm{OT}_{1.0\mathrm{s}}$ and $\mathrm{OT}_{1.5\mathrm{s}}$ measure timely responses within 1.0 and 1.5 seconds, respectively.

\paragraph{Evaluated Events.}
We evaluate local prompt updates after the initial prompt, excluding updates that reset the local prompt to the shared environment prompt.
For event $j$ in video $c$, let $a_{cj}$ be its prompt-update time and $b_{cj}$ the next prompt-update time, with $b_{cj}=+\infty$ for the final update.
Let $T_c=(F_c-1)/f_c$ be the timestamp of the last valid frame, where $F_c$ and $f_c$ are the video frame count and frame rate.
The event is evaluated within $\mathcal{I}_{cj}=[a_{cj},b_{cj})\cap[0,T_c]$.
Thus, a new prompt update ends the preceding event interval, including when the new prompt resets to the environment description.
We include an event only if $a_{cj}<T_c$ and its interval contains at least one valid video frame.

Let $J_c$ denote the number of evaluated events in video $c$, and let $\mathcal{C}_{+}=\{c:J_c>0\}$ denote the videos containing at least one such event.
All methods use the same event definitions and evaluation videos.
An event remains in the denominator even if the model does not respond.

\paragraph{Event Adherence (EA).}
EA measures how completely the model performs the requested events.
Using visual evidence within $\mathcal{I}_{cj}$, the evaluator assigns
\begin{equation}
s_{cj}=
\begin{cases}
1, & \text{the requested event is clearly completed},\\
0.5, & \text{the event is recognizable but incomplete},\\
0, & \text{the event is absent or incorrect}.
\end{cases}
\label{eq:event_completion_score}
\end{equation}
Matching scene appearance, unrelated camera motion, or unchanged content from before the update alone does not count as event completion.
We average the event scores within each video and then average equally across videos:
\begin{equation}
\mathrm{EA}
=\frac{100}{|\mathcal{C}_{+}|}
\sum_{c\in\mathcal{C}_{+}}\frac{1}{J_c}
\sum_{j=1}^{J_c}s_{cj}.
\label{eq:event_adherence}
\end{equation}

\begin{samepage}
\paragraph{On-Time Accuracy (OT).}
OT measures whether the model responds correctly within $\tau$ seconds of a prompt update.
Let $r_{cj}$ be the timestamp of the first inspected frame within $\mathcal{I}_{cj}$ showing clear new evidence of the requested event.
If no such response is observed, we set $r_{cj}=+\infty$.
The event receives timing credit only when the response occurs within both its active interval and the time threshold:
\begin{equation}
\begin{aligned}
q_{cj}^{(\tau)}
&=\mathbf{1}\!\left[
r_{cj}\in\mathcal{I}_{cj}\;\land\;
0\leq r_{cj}-a_{cj}\leq\tau
\right],\\[3pt]
\mathrm{OT}_{\tau}
&=\frac{100}{|\mathcal{C}_{+}|}
\sum_{c\in\mathcal{C}_{+}}\frac{1}{J_c}
\sum_{j=1}^{J_c}q_{cj}^{(\tau)},
\qquad \tau\in\{1.0\mathrm{s},1.5\mathrm{s}\},
\end{aligned}
\label{eq:event_on_time}
\end{equation}
where $\mathbf{1}[\cdot]$ is the indicator function.
We compute both OT scores from the same recorded response timestamps, measured in generated-video time rather than inference time.
A response after the next prompt update receives no credit for the preceding event, even if it is within $\tau$ seconds of that event's start.
A recognizable partial response can receive OT credit without completing the event, so OT can exceed EA.
Both metrics are reported on a 0--100 scale, with higher scores indicating better performance.
Events have equal weight within each video, and videos have equal weight regardless of their number of events.
Videos with $J_c=0$ are excluded rather than assigned zero scores; if $\mathcal{C}_{+}$ is empty, both metrics are reported as N/A.

\end{samepage}

\subsection{User Study}
\label{app:user_study}
We conduct a pairwise user study on InBench to evaluate visual quality, interactive following, and motion clarity.

\paragraph{Pairwise Comparison.}
Each trial compares our model with one of five baselines: HY-WorldPlay, JoyAI-Echo-WM, LingBot-World-Fast, Matrix-Game~3.0, and SANA-WM-Streaming.
The two videos use the same initial image and source control sequence and are displayed side by side at the same size using the 480p evaluation outputs.
They are labeled only as \emph{Video A} and \emph{Video B}, with the left/right position of our model randomized independently for each trial.
Model identities and numerical evaluation scores are hidden from participants.

\paragraph{Case Allocation.}
The study uses the full InBench case pool, comprising 96 cases: 64 roaming cases and 32 interactive cases.
Each session contains 20 pairs of videos (40 videos in total), corresponding to 10 interactive cases and 10 roaming cases, with no case repeated within a session.
Each baseline appears in four trials per session.
We shuffle the baseline assignments and select cases within each subset, prioritizing case--baseline pairs with fewer previous assignments.
The two subsets are presented in separate blocks, and no case is repeated within a session.

\paragraph{Evaluation Questions.}
For each applicable question, participants select \emph{Video A is better}, \emph{They are the same}, or \emph{Video B is better}.
The three evaluation dimensions are:
\begin{itemize}
  \item \textbf{Visual quality (both subsets).} Participants compare overall appearance, temporal stability, natural motion, and visible artifacts.
  \item \textbf{Interactive following (interactive subset).} Participants compare how well the videos perform the requested events as the corresponding local prompts become active.
  \item \textbf{Motion clarity (roaming subset).} Participants compare scene detail during camera movement, focusing on blur, smearing, and ghosting. They judge frames during movement and select a tie when clarity is similar or no viewpoint change is visible.
\end{itemize}
Each trial therefore collects a visual-quality judgment and one judgment specific to its subset.
\Cref{fig:user_study_interface} illustrates the evaluation interface.

\clearpage
\begin{figure}[H]
\centering
\includegraphics[width=\linewidth]{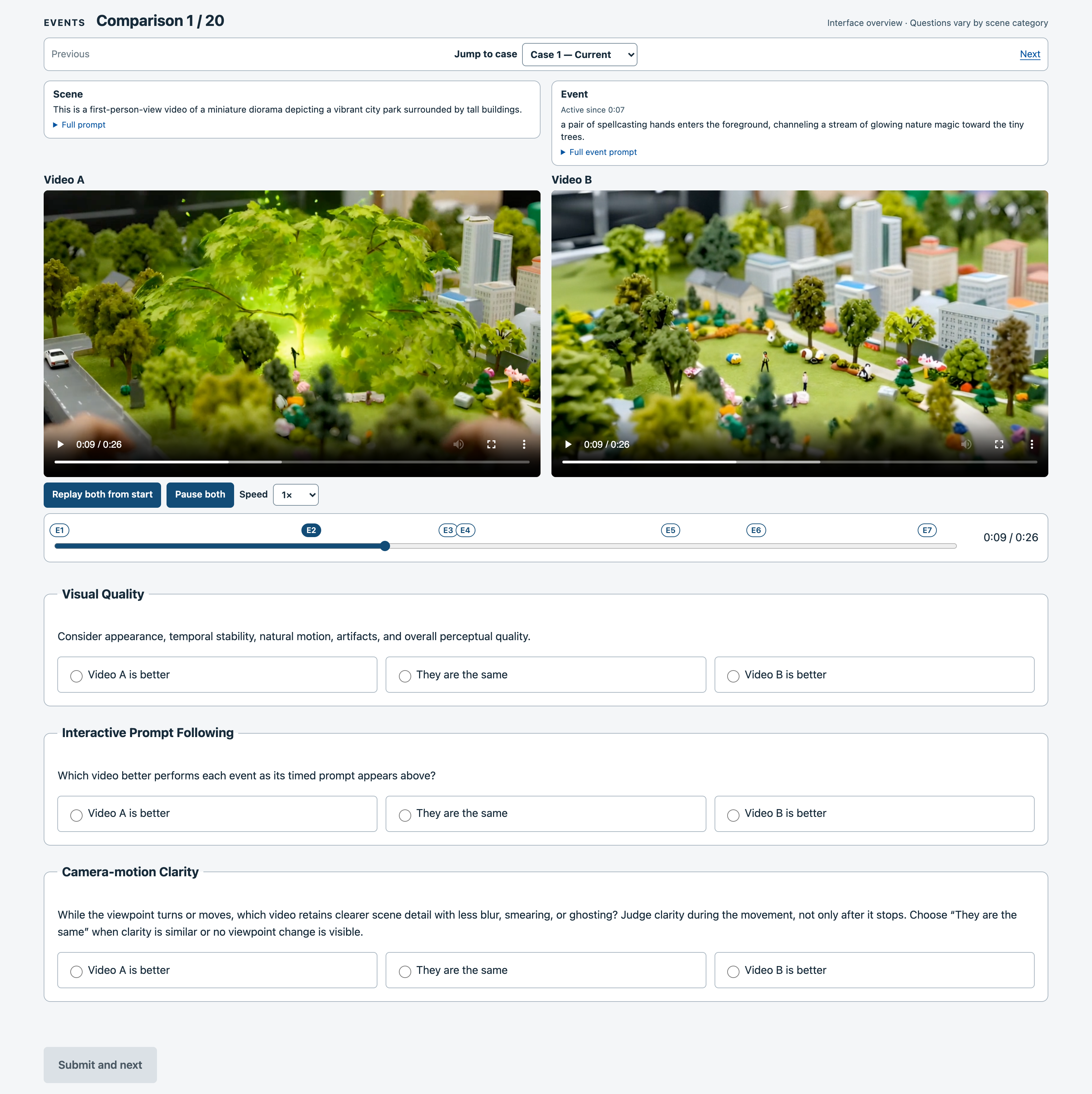}
\caption{\textbf{User-Study interface.}
}
\label{fig:user_study_interface}
\end{figure}

\paragraph{Preference Aggregation.}
We map the Video A/B choices back to model identities.
For baseline $b$ and question $q$, let $W_{bq}$, $T_{bq}$, and $L_{bq}$ denote the numbers of answers preferring our model, a tie, and the baseline, respectively.
With $N_{bq}=W_{bq}+T_{bq}+L_{bq}$, the reported preference rates are
\begin{equation}
\left(P^{\mathrm{ours}}_{bq},P^{\mathrm{tie}}_{bq},P^{\mathrm{base}}_{bq}\right)
=\frac{100}{N_{bq}}\left(W_{bq},T_{bq},L_{bq}\right).
\label{eq:user_study_preference}
\end{equation}
Each submitted answer has equal weight within its baseline--question group, and ties are retained as a separate outcome.
We include completed trials from partially completed sessions and exclude unanswered questions.
Visual-quality preferences combine judgments from both subsets, while interactive-following and motion-clarity preferences use only judgments from the interactive and roaming subsets, respectively.

\clearpage
\section{Additional Experimental Results}
\label{app:experimental_results}
\label{app:quantitative}
\label{app:qualitative}

\subsection{Comparison with State-of-the-Art Methods}
\label{app:qual_comparison}

\paragraph{Results on WBench.}
\label{app:wbench}

\Cref{tab:wbench} reports the complete results on the Navi-158 subset of WBench~\cite{ying2026wbench}, covering generation quality, setting adherence, and navigation performance.
Our model achieves the highest average Quality and Setting scores of 78.5 and 80.3, respectively.
It also obtains the highest AQ, IQ, HPSv3, and subject-adherence scores, indicating strong visual quality and adherence to the specified subjects.
The average Interaction score is 84.3.
We attribute this gap partly to a mismatch between the global prompts used in WBench and the chunk-wise prompt conditioning used in our cross-attention.
This interpretation is supported by the cross-attention ablation in the main paper, where chunk-wise prompts reduce all three camera errors compared with global prompts.

\begin{table}[H]
\centering
\caption{\textbf{Quantitative comparison on WBench Navi-158.}
The \textit{Setting} group measures scene- and subject-level adherence, while the \textit{Interaction} group reports navigation accuracy (Acc) and consistency (Cons).
Avg denotes the mean within each group.
Darker highlights indicate higher ranks among the three best averages.
}
\label{tab:wbench}
\begingroup
\fontsize{8}{10}\selectfont
\setlength{\tabcolsep}{0.5pt}
\renewcommand{\arraystretch}{1.15}
\begin{tabularx}{\linewidth}{l*{12}{N}}
\toprule
\multicolumn{1}{c}{\multirow{2}{*}{\textbf{Method}}} &
\multicolumn{6}{c}{\textbf{Quality} $\boldsymbol{\uparrow}$} &
\multicolumn{3}{c}{\textbf{Setting} $\boldsymbol{\uparrow}$} &
\multicolumn{3}{c}{\textbf{Interaction} $\boldsymbol{\uparrow}$} \\
\cmidrule(lr){2-7}\cmidrule(lr){8-10}\cmidrule(l){11-13}
& \textbf{AQ} & \textbf{IQ} & \textbf{TFlk} &
\textbf{DD} & \textbf{HPSv3} & \textbf{Avg} &
\textbf{Scene} & \textbf{Subject} & \textbf{Avg} &
\textbf{Acc} & \textbf{Cons} & \textbf{Avg} \\
\midrule
HY-WorldPlay
& 63.8 & 67.0 & 93.7 & 66.5 & 71.8 & 72.6
& 56.5 & 89.7 & 73.1
& 83.0 & 91.5 & \cellcolor{tableRank1}\textbf{87.2} \\
JoyAI-Echo-WM
& 62.8 & 65.9 & 92.2 & 94.9 & 64.4
& \cellcolor{tableRank2}76.0
& 60.6 & 89.0 & 74.8
& 85.3 & 87.7 & \cellcolor{tableRank2}86.5 \\
Matrix-Game 3.0
& 57.7 & 64.9 & 91.1 & 94.3 & 56.1 & 72.8
& 50.0 & 80.7 & 65.3
& 82.2 & 82.7 & 82.5 \\
SANA-WM-Streaming
& 60.6 & 62.8 & 94.7 & 95.6 & 61.2
& 75.0
& 62.3 & 87.7 & \cellcolor{tableRank3}75.0
& 82.3 & 80.6 & 81.5 \\
LingBot-World-Fast
& 60.1 & 63.1 & 92.4 & 98.7 & 61.9
& \cellcolor{tableRank3}75.3
& 70.2 & 90.1 & \cellcolor{tableRank2}80.2
& 83.2 & 79.6 & 81.4 \\
\textbf{\method{}~(Ours)}
& 64.0 & 68.1 & 91.2 & 93.7 & 75.8
& \cellcolor{tableRank1}\textbf{78.5}
& 69.2 & 91.5 & \cellcolor{tableRank1}\textbf{80.3}
& 81.7 & 86.9 & \cellcolor{tableRank3}84.3 \\
\bottomrule
\end{tabularx}
\endgroup
\end{table}

\paragraph{Results on SANA-WM-Bench.}
\label{app:sanawm}

\Cref{tab:sanawm_quality} provides the individual quality metrics on the Simple and Hard 60-second trajectory splits of SANA-WM-Bench~\cite{zhu2026sanawm}, supplementing the average Quality scores reported in the main paper.
Our model achieves the highest average scores of 74.8 and 74.7 on the two splits, respectively.
On the Simple split, it also obtains the highest IQ and HPSv3 scores of 71.1 and 60.6.
On the Hard split, it matches the highest DD score of 97.5 and achieves an HPSv3 score of 58.6, close to the best score of 58.8.

\begin{table}[H]
\centering
\caption{\textbf{Detailed quality results on SANA-WM-Bench.}
Quality metrics are evaluated on 60-second videos.
Avg denotes their arithmetic mean; higher scores are better.
Darker highlights indicate higher ranks among the three best averages within each split.
Metric definitions are provided in \cref{app:quality_camera_metrics}.}
\label{tab:sanawm_quality}
\begingroup
\fontsize{8}{10}\selectfont
\setlength{\tabcolsep}{2pt}
\renewcommand{\arraystretch}{1.15}
\begin{tabularx}{\linewidth}{l*{6}{N}}
\toprule
\multicolumn{1}{c}{\multirow{2}{*}{\textbf{Method}}} & \multicolumn{6}{c}{\textbf{Quality} $\boldsymbol{\uparrow}$} \\
\cmidrule(l){2-7}
& \textbf{AQ} & \textbf{IQ} & \textbf{TFlk} & \textbf{DD} & \textbf{HPSv3} & \textbf{Avg} \\
\midrule
\multicolumn{7}{@{}l}{\textit{Simple Trajectory Split (60s)}} \\
\midrule
HY-WorldPlay & 56.7 & 69.3 & 97.9 & 25.0 & 59.7 & 61.7 \\
JoyAI-Echo-WM & 55.4 & 69.5 & 96.3 & 66.3 & 59.6 & 69.4 \\
Matrix-Game 3.0 & 50.4 & 68.4 & 93.1 & 97.5 & 42.8 & \cellcolor{tableRank2}70.4 \\
SANA-WM-Streaming & 51.1 & 63.5 & 97.7 & 90.0 & 46.1 & 69.7 \\
LingBot-World-Fast & 53.1 & 61.2 & 96.4 & 92.5 & 47.2 & \cellcolor{tableRank3}70.1 \\
\textbf{\method{}~(Ours)} & 54.8 & 71.1 & 94.9 & 92.5 & 60.6 & \cellcolor{tableRank1}\textbf{74.8} \\
\midrule
\multicolumn{7}{@{}l}{\textit{Hard Trajectory Split (60s)}} \\
\midrule
HY-WorldPlay & 55.9 & 69.5 & 97.1 & 46.3 & 58.8 & 65.5 \\
JoyAI-Echo-WM & 54.3 & 69.3 & 95.5 & 88.8 & 56.9 & \cellcolor{tableRank2}72.9 \\
Matrix-Game 3.0 & 50.6 & 68.5 & 93.3 & 96.3 & 42.4 & 70.2 \\
SANA-WM-Streaming & 50.7 & 63.4 & 97.0 & 97.5 & 47.2 & \cellcolor{tableRank3}71.1 \\
LingBot-World-Fast & 52.2 & 62.9 & 96.1 & 92.5 & 45.8 & 69.9 \\
\textbf{\method{}~(Ours)} & 54.1 & 69.3 & 94.3 & 97.5 & 58.6 & \cellcolor{tableRank1}\textbf{74.7} \\
\bottomrule
\end{tabularx}
\endgroup
\end{table}

\clearpage
\paragraph{Qualitative Comparison on InBench.}
\Crefrange{fig:inbench_comparison_01}{fig:inbench_comparison_09} compare our interactive video world model with HY-WorldPlay, JoyAI-Echo-WM, Matrix-Game~3.0, SANA-WM-Streaming, and LingBot-World-Fast on 18 selected InBench cases.
For each case, we show five frames at the same generated-video timestamps across all methods, with summaries of the active event instructions above the columns.
For roaming cases, the five frames are sampled uniformly over the full video duration.
The examples include environmental changes, character actions, and exploration across diverse scenes.
In the portal, rainbow, and ocean-view room examples, our model responds to the requested events while retaining recognizable scene content.
The roaming examples further illustrate subject and scene consistency during camera motion.

\begin{figure}[H]
\centering
\includegraphics[page=1,width=0.90\linewidth]{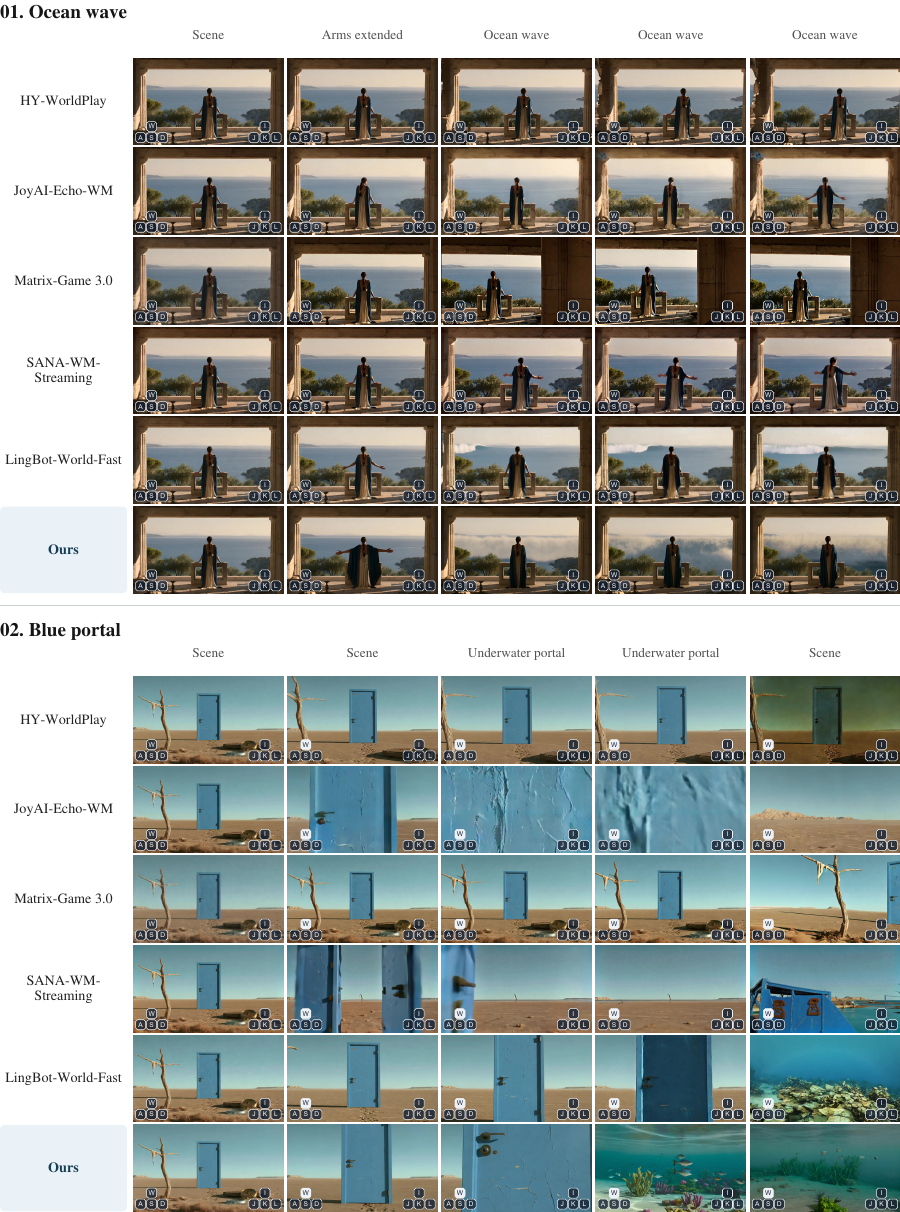}
\caption{\textbf{Qualitative comparisons with state-of-the-art methods (1/9).}
Each group shows one case, with six methods evaluated at matched generated-video timestamps.
Column headers summarize the active event instructions; ``Scene'' denotes environment-only conditioning.
Pressed WASD/IJKL keys are shown with a light background according to the recorded keyboard inputs.}
\label{fig:inbench_comparison_01}
\end{figure}

\clearpage
\begin{figure}[H]
\centering
\includegraphics[page=2,width=\linewidth]{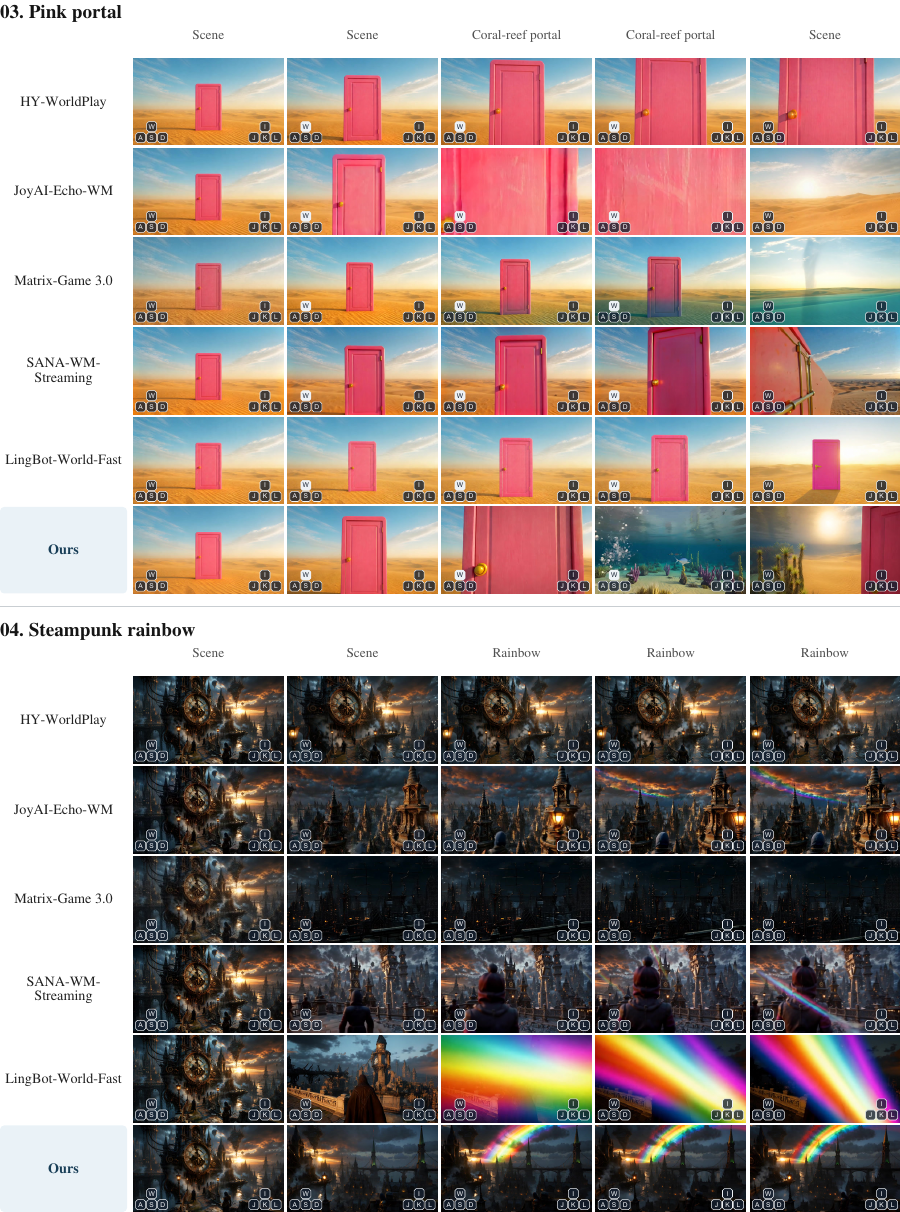}
\caption{\textbf{Qualitative comparisons with state-of-the-art methods (2/9).}
Each group shows one case, with six methods evaluated at matched generated-video timestamps.
Column headers summarize the active event instructions; ``Scene'' denotes environment-only conditioning.
Pressed WASD/IJKL keys are shown with a light background according to the recorded keyboard inputs.}
\label{fig:inbench_comparison_02}
\end{figure}

\clearpage
\begin{figure}[H]
\centering
\includegraphics[page=3,width=\linewidth]{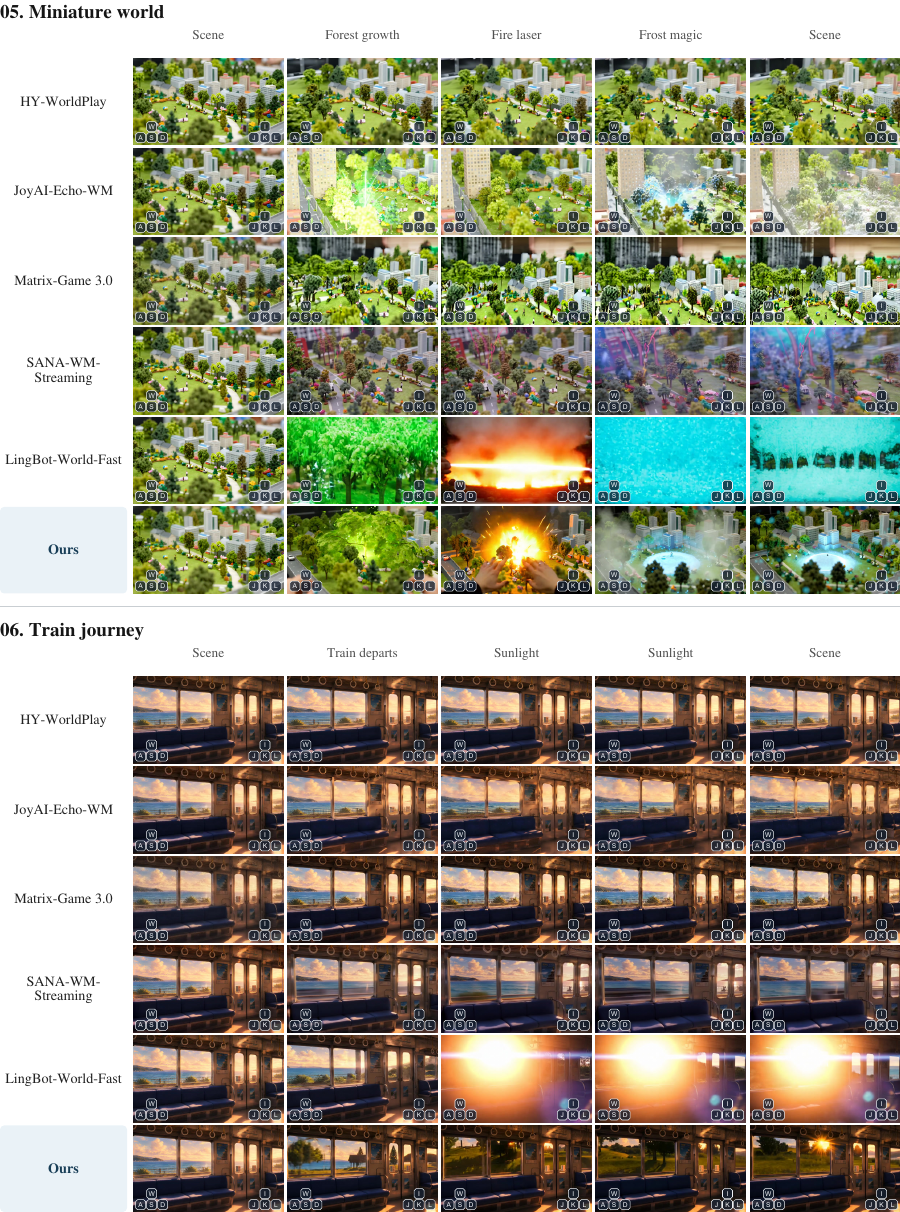}
\caption{\textbf{Qualitative comparisons with state-of-the-art methods (3/9).}
Each group shows one case, with six methods evaluated at matched generated-video timestamps.
Column headers summarize the active event instructions; ``Scene'' denotes environment-only conditioning.
Pressed WASD/IJKL keys are shown with a light background according to the recorded keyboard inputs.}
\label{fig:inbench_comparison_03}
\end{figure}

\clearpage
\begin{figure}[H]
\centering
\includegraphics[page=4,width=\linewidth]{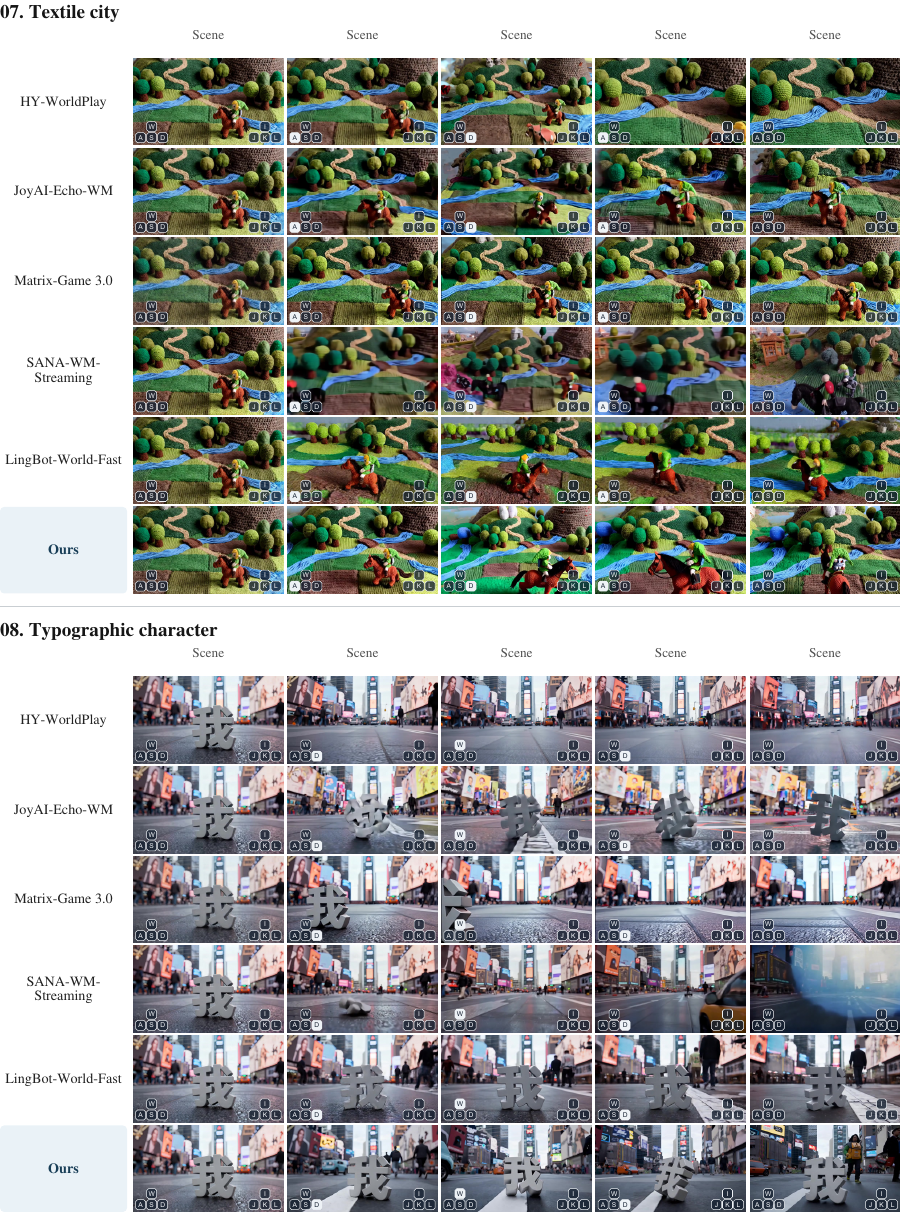}
\caption{\textbf{Qualitative comparisons with state-of-the-art methods (4/9).}
Each group shows one case, with six methods evaluated at matched generated-video timestamps.
Column headers summarize the active event instructions; ``Scene'' denotes environment-only conditioning.
Pressed WASD/IJKL keys are shown with a light background according to the recorded keyboard inputs.}
\label{fig:inbench_comparison_04}
\end{figure}

\clearpage
\begin{figure}[H]
\centering
\includegraphics[page=5,width=\linewidth]{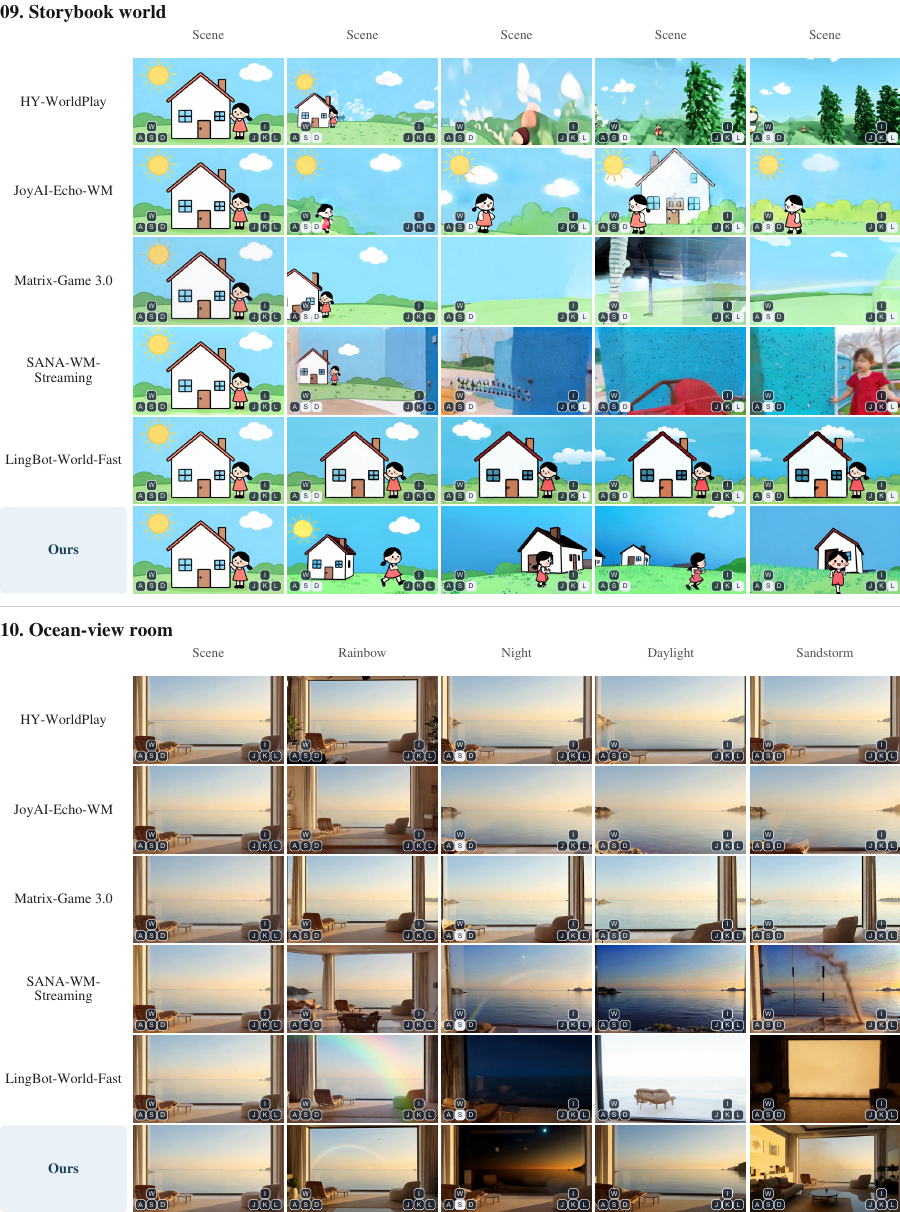}
\caption{\textbf{Qualitative comparisons with state-of-the-art methods (5/9).}
Each group shows one case, with six methods evaluated at matched generated-video timestamps.
Column headers summarize the active event instructions; ``Scene'' denotes environment-only conditioning.
Pressed WASD/IJKL keys are shown with a light background according to the recorded keyboard inputs.}
\label{fig:inbench_comparison_05}
\end{figure}

\clearpage
\begin{figure}[H]
\centering
\includegraphics[page=6,width=\linewidth]{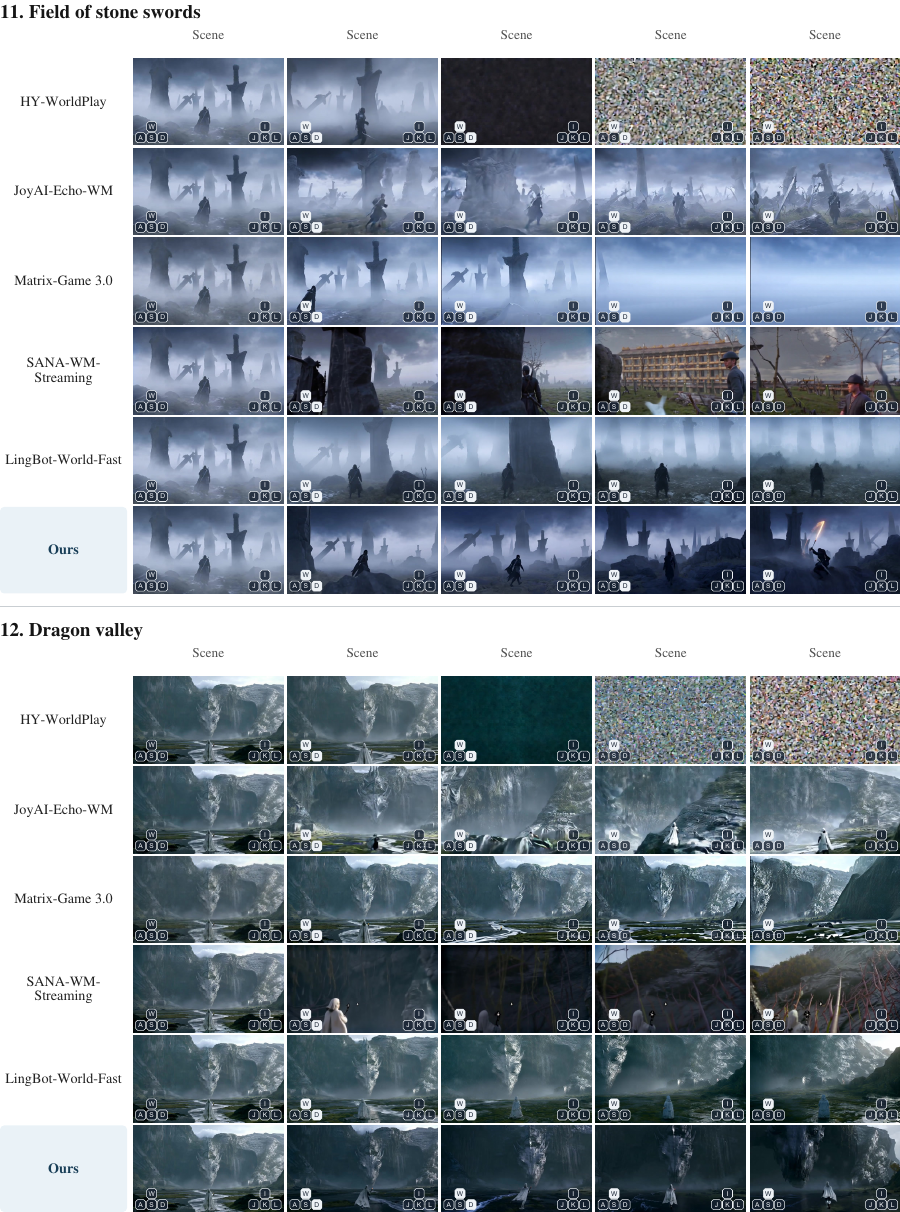}
\caption{\textbf{Qualitative comparisons with state-of-the-art methods (6/9).}
Each group shows one case, with six methods evaluated at matched generated-video timestamps.
Column headers summarize the active event instructions; ``Scene'' denotes environment-only conditioning.
Pressed WASD/IJKL keys are shown with a light background according to the recorded keyboard inputs.}
\label{fig:inbench_comparison_06}
\end{figure}

\clearpage
\begin{figure}[H]
\centering
\includegraphics[page=7,width=\linewidth]{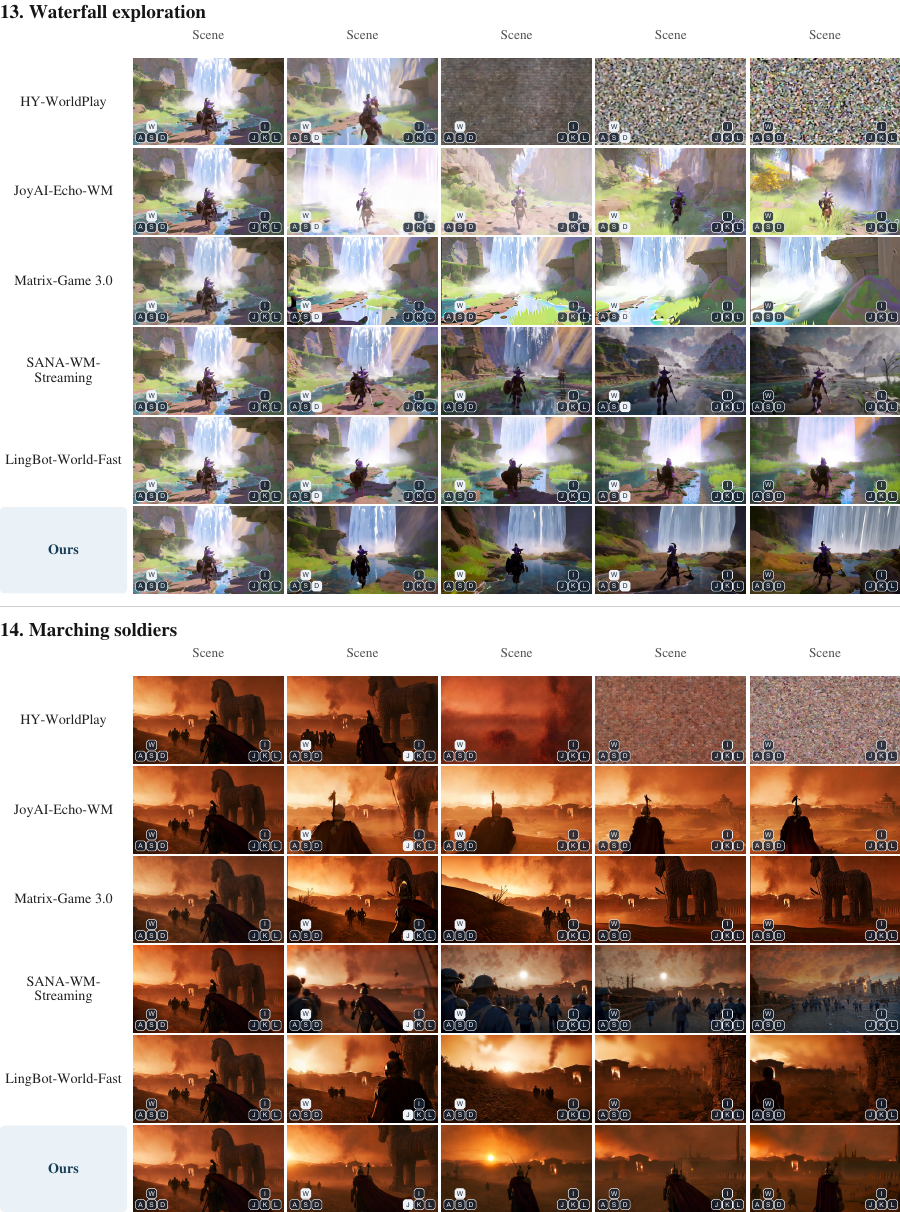}
\caption{\textbf{Qualitative comparisons with state-of-the-art methods (7/9).}
Each group shows one case, with six methods evaluated at matched generated-video timestamps.
Column headers summarize the active event instructions; ``Scene'' denotes environment-only conditioning.
Pressed WASD/IJKL keys are shown with a light background according to the recorded keyboard inputs.}
\label{fig:inbench_comparison_07}
\end{figure}

\clearpage
\begin{figure}[H]
\centering
\includegraphics[page=8,width=\linewidth]{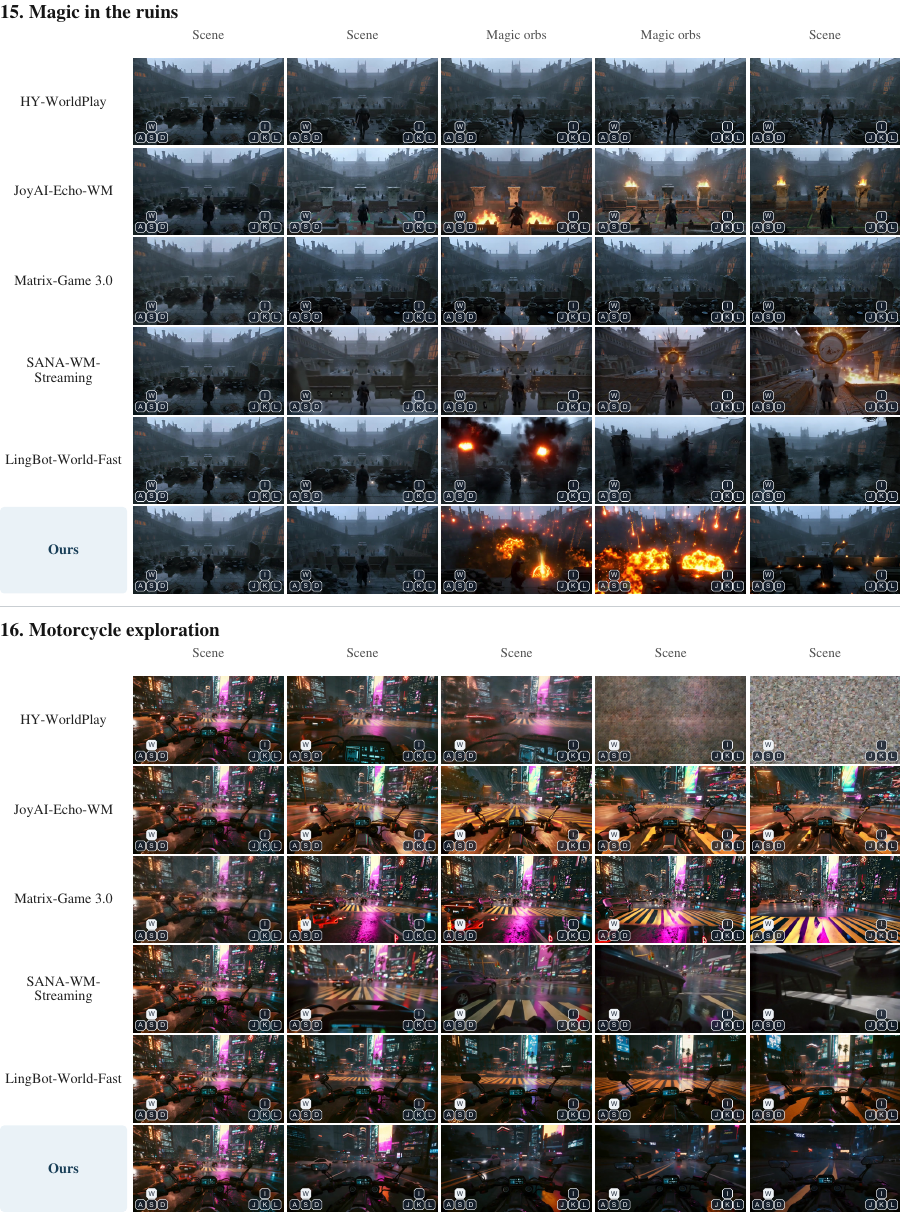}
\caption{\textbf{Qualitative comparisons with state-of-the-art methods (8/9).}
Each group shows one case, with six methods evaluated at matched generated-video timestamps.
Column headers summarize the active event instructions; ``Scene'' denotes environment-only conditioning.
Pressed WASD/IJKL keys are shown with a light background according to the recorded keyboard inputs.}
\label{fig:inbench_comparison_08}
\end{figure}

\clearpage
\begin{figure}[H]
\centering
\includegraphics[page=9,width=\linewidth]{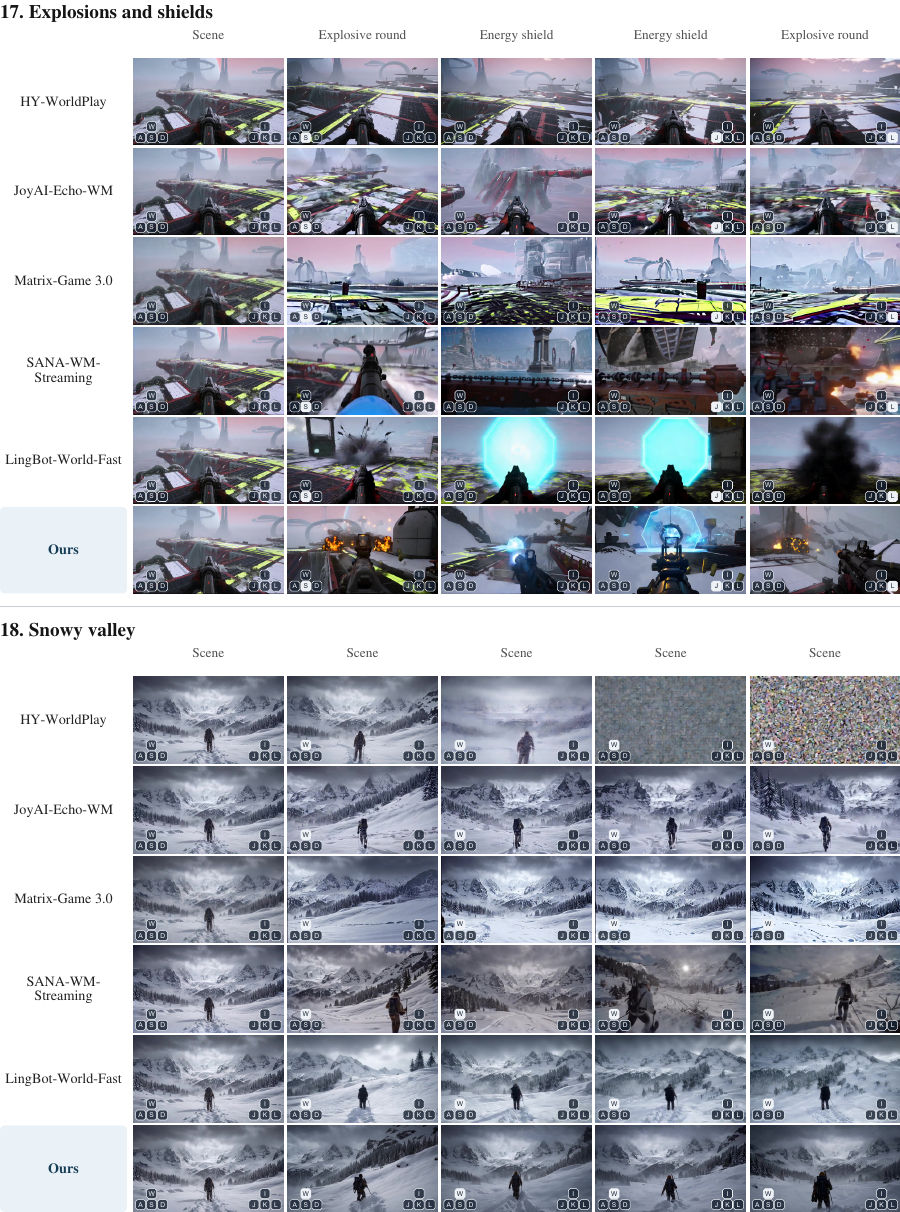}
\caption{\textbf{Qualitative comparisons with state-of-the-art methods (9/9).}
Each group shows one case, with six methods evaluated at matched generated-video timestamps.
Column headers summarize the active event instructions; ``Scene'' denotes environment-only conditioning.
Pressed WASD/IJKL keys are shown with a light background according to the recorded keyboard inputs.}
\label{fig:inbench_comparison_09}
\end{figure}

\clearpage

\subsection{Causal Pre-Training}
\label{app:qual_pretrain}
\Cref{fig:pretrain} presents 60-second rollouts from our causal checkpoint before few-step post-training.
The eagle and fire-extinguisher comparisons show retained subject appearance and scene details at late timestamps, while the camera-controlled example further illustrates generation quality during exploration.

\begin{figure}[H]
\centering
\includegraphics[width=0.90\linewidth]{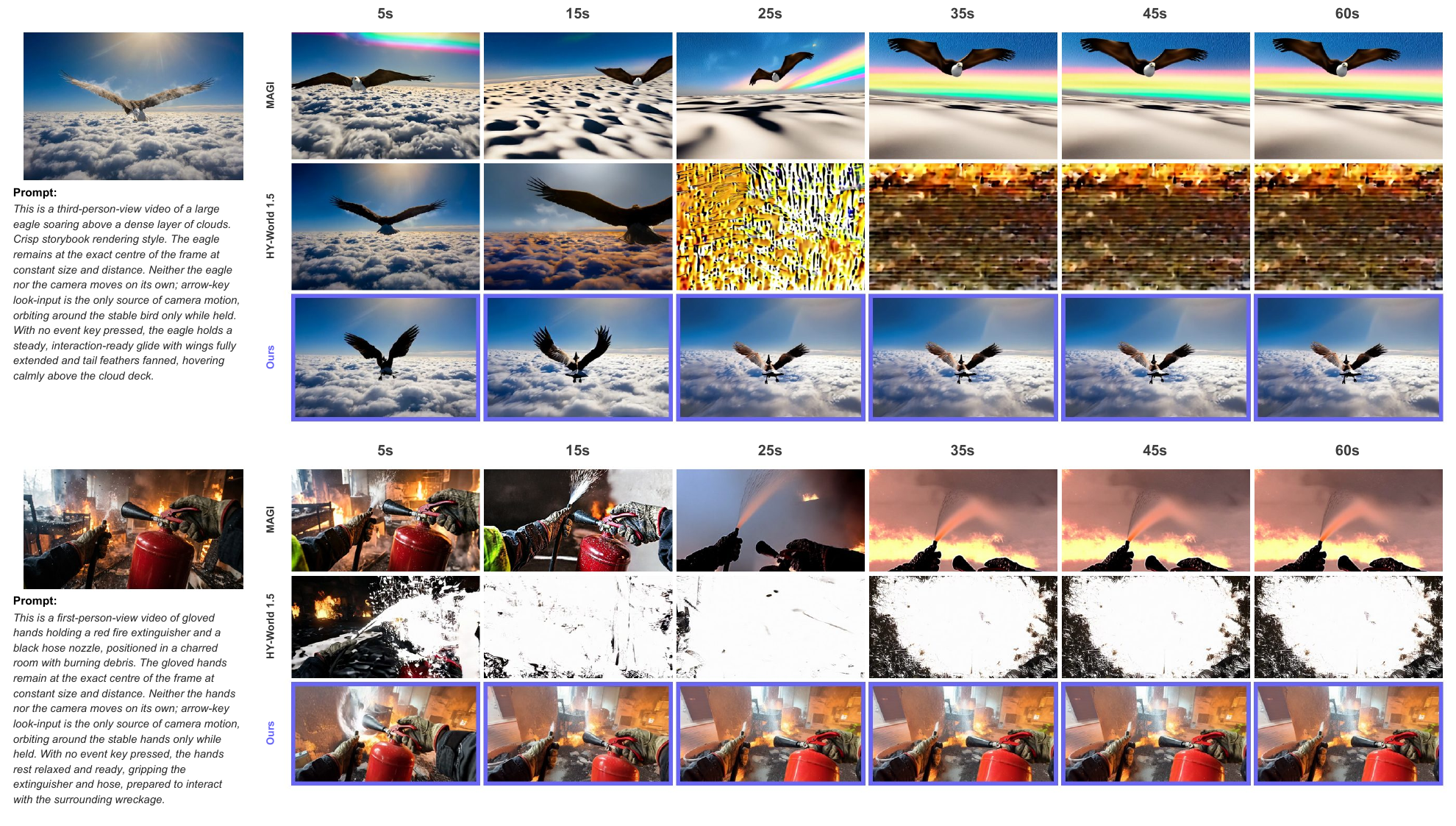}
\par\vspace{5pt}
\includegraphics[width=0.90\linewidth]{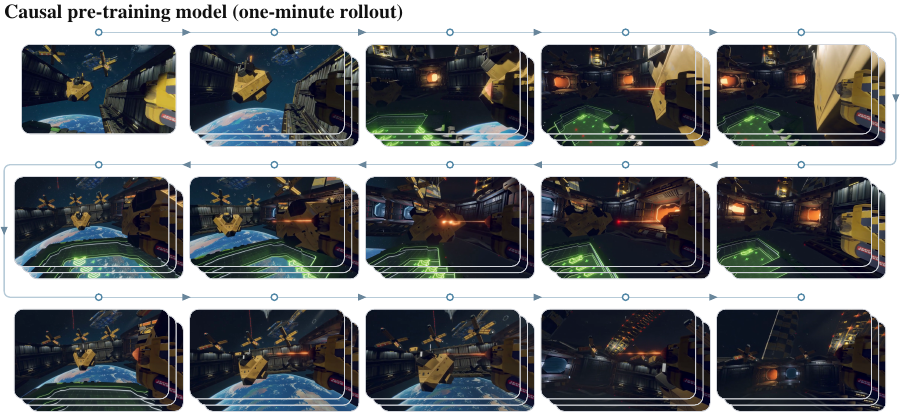}
\caption{\textbf{Qualitative results of causal pre-training.}
Top: comparisons with MAGI and HY-World~1.5 over 60-second rollouts, with the initial image and full prompt above each group.
Bottom: a one-minute camera-controlled rollout from our causal checkpoint.
All our results are generated before few-step post-training.}
\label{fig:pretrain}
\label{fig:causal_pretrain_rollout}
\end{figure}
\clearpage

\subsection{Long-Horizon Generation}
\label{app:long_horizon}
\label{app:long_horizon_quality}
\Cref{fig:long_horizon} shows a one-hour rollout from our interactive video world model, traversing diverse environments from canals and city streets to indoor spaces and orbital views.
We further evaluate the performance of one-hour rollouts, dividing each video into twelve consecutive five-minute intervals.
For each interval, we compute the Quality score defined in \cref{eq:supp_quality_average} and report the mean and standard deviation across rollouts.
The mean remains between 67.3 and 71.6, with a score of 69.3 in the final interval compared with 71.0 in the first, indicating relatively stable generation quality over one hour.

\begin{figure}[H]
\centering
\includegraphics[width=\linewidth]{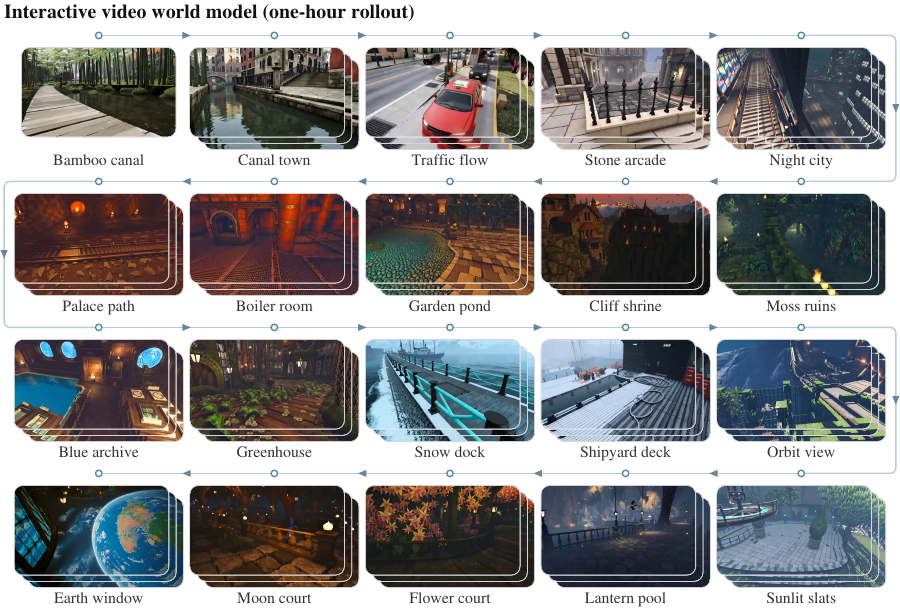}
\par\vspace{6pt}
\includegraphics[width=\linewidth]{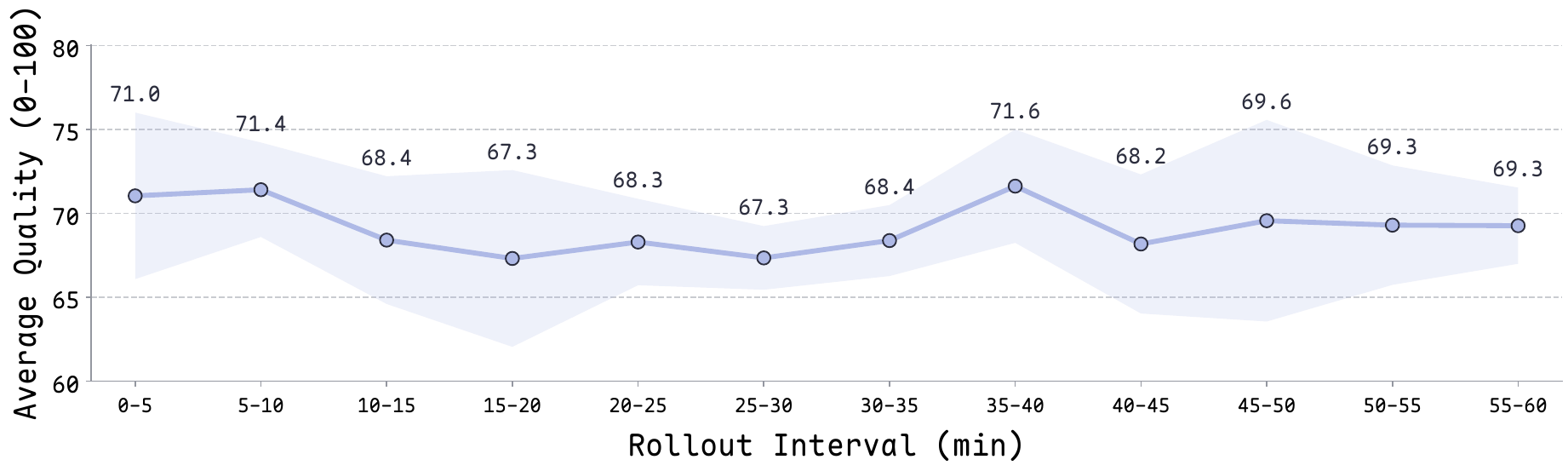}
\caption{\textbf{One-Hour autoregressive generation.}
Top: representative frames from a one-hour rollout of our interactive video world model.
Bottom: mean Quality scores across one-hour rollouts, evaluated in consecutive five-minute intervals.
The shaded region indicates the standard deviation across rollouts.
Scores are on a 0--100 scale; higher is better.}
\label{fig:long_horizon}
\label{fig:long_rollout_quality}
\end{figure}

\clearpage
\subsection{Additional Interactive Results}
\label{app:qual_interaction}
\Cref{fig:result1,fig:result2} present additional examples of online prompt interaction and camera-controlled exploration using our interactive video world model, illustrating changes in character actions and scene content in response to online prompts.

\begin{figure}[H]
\centering
\includegraphics[width=0.90\linewidth]{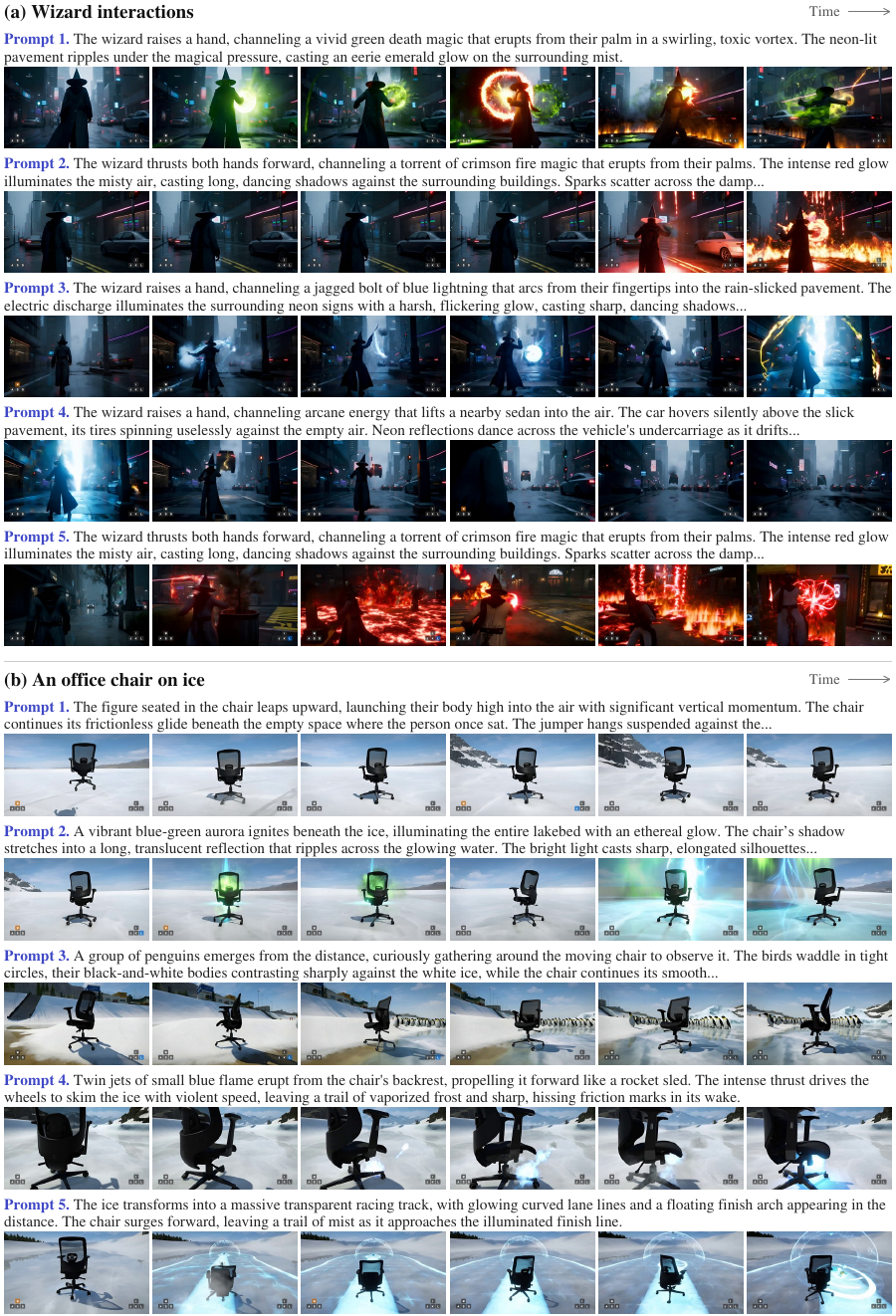}
\caption{\textbf{Additional interactive results (1/2).}
(a) Magical effects around a wizard. (b) An office chair on ice, with aurora effects and the appearance of penguins.
Each row shows six frames from a prompt interval in temporal order, with the prompt excerpt above.
Rows follow the rollout sequence within each group.}
\label{fig:result1}
\end{figure}

\clearpage
\begin{figure}[H]
\centering
\includegraphics[width=0.90\linewidth]{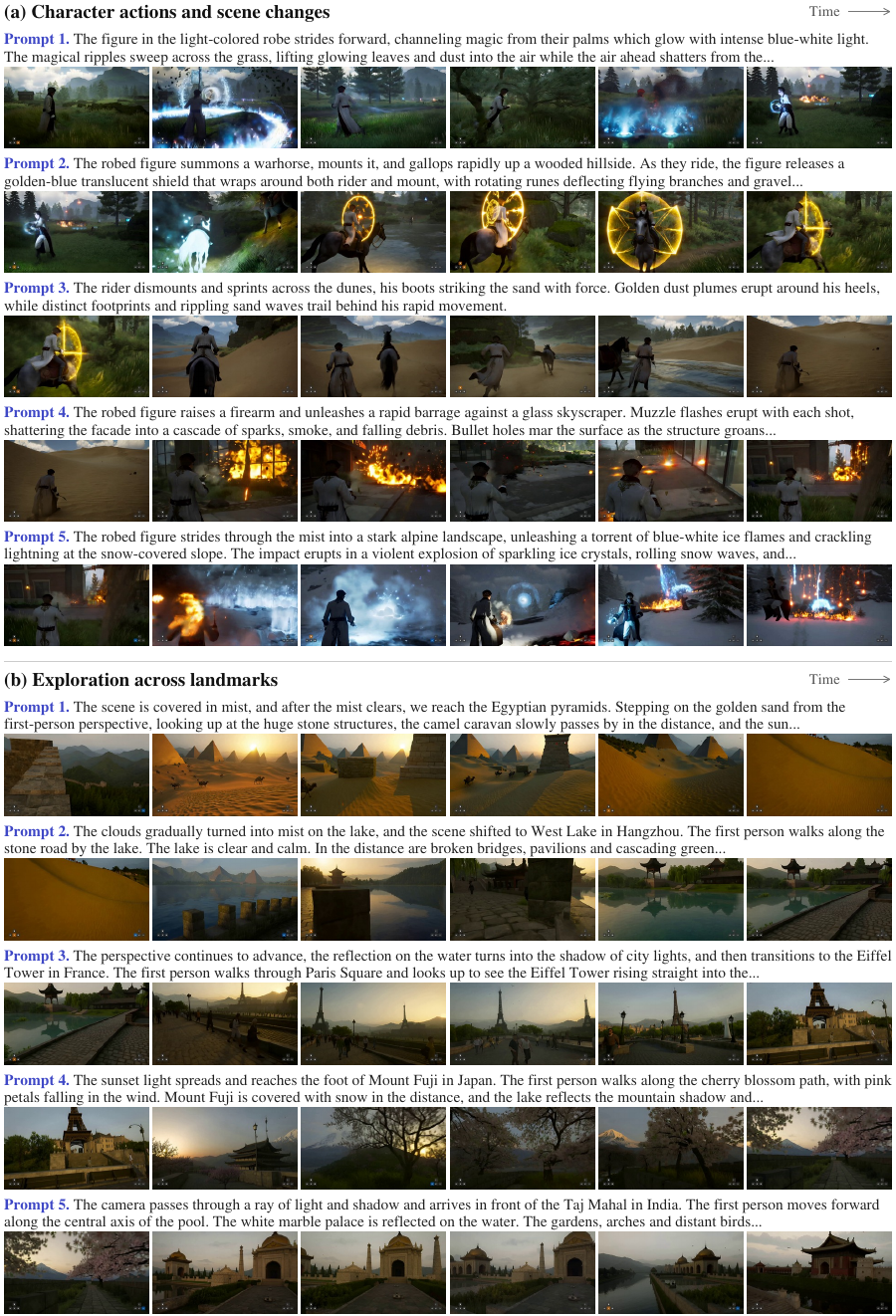}
\caption{\textbf{Additional interactive results (2/2).}
(a) Character actions including casting magic, riding a horse, and running.
(b) First-person exploration across the Egyptian pyramids, West Lake, the Eiffel Tower, Mount Fuji, and the Taj Mahal.
Each row shows a prompt excerpt and six frames in temporal order, following the layout of \cref{fig:result1}.}
\label{fig:result2}
\end{figure}

\clearpage

  \stopcontents[appendix]
\fi

\clearpage
\bibliographystyle{iclr2027_conference}
\bibliography{references}
\end{document}